\documentclass{article}

\usepackage[classic]{sotalab}
\usepackage[T1]{fontenc}
\usepackage[utf8]{inputenc}
\usepackage{natbib}
\usepackage{hyperref}
\usepackage{url}
\usepackage{doi}
\usepackage{cleveref}
\usepackage{tabularx}
\usepackage{multirow}
\usepackage{booktabs}
\usepackage{array}
\usepackage{graphicx}
\usepackage{caption}
\usepackage{float}
\usepackage{placeins}
\usepackage{authblk}

\newcolumntype{Y}{>{\raggedright\arraybackslash}X}
\newcommand{\passone}{\textnormal{pass@1}}
\newcommand{\passatthree}{\textnormal{pass@3}}

\newcommand{\passthree}{\textnormal{pass\textasciicircum{}3}}

\title{MathVis: Can AI Reason through Drawing? \\
  A Benchmark for \titleaccent{Mathematical Figure Generation}}
\renewcommand{\shorttitle}{MathVis}
\renewcommand{\headeright}{Preprint}
\renewcommand{\undertitle}{Preprint}
\venue{SoTALab \textperiodcentered{} Frontier Data Lab}
\reportid{MATHVIS-2026-09}
\date{September 2026}

\author[1]{Jingjing Wei}
\author[2]{Hao Zhao}
\author[3]{Xinke Li}
\author[4]{Linxi Yu}
\author[1]{Xiaosong Gao}
\author[5]{Shanshan Xiao}

\affil[1]{Zhihu (zhihu.com)}
\affil[2]{Tsinghua University (tsinghua.edu.cn)}
\affil[3]{City University of Hong Kong (cityu.edu.hk)}
\affil[4]{CIS, Cornell University (cornell.edu)}
\affil[5]{Tsinghua University (tsinghua.edu.cn)}

\hypersetup{
  pdftitle={MathVis: Can AI Reason through Drawing? A Benchmark for Mathematical Figure Generation},
  pdfauthor={Jingjing Wei, Hao Zhao, Xinke Li, Linxi Yu, Xiaosong Gao, Shanshan Xiao},
  pdfsubject={cs.AI, cs.CL, cs.CV},
  pdfkeywords={evaluation benchmark, MathVis, large language models, advanced mathematical figure generation, executable graphics, TikZ/SVG},
}

\begin{document}
\maketitle
\begin{abstract}
This paper introduces MathVis, a benchmark designed to evaluate LLMs' ability to visualize frontier mathematical problems and perform diagram-assisted reasoning. The benchmark consists of 512 tasks, each comprising a mathematical statement, a reference diagram, executable TikZ/SVG code, a Gold Intermediate Representation (IR), atomic rubrics, and error codes. Through rigorous evaluation, GPT-6-astra, Claude Fable 5, and Gemini 3.1 Pro achieve Pass@1 scores of merely $31.25\%$, $26.56\%$, and $22.85\%$, respectively. We found that although the tasks require visualization, LLMs still reason primarily through formal languages and symbols, rarely relying on visual diagrams. Furthermore, while the solutions generated by LLMs may contain all the necessary elements, they are often visually confusing to humans. Ultimately, this benchmark aims to visualize mathematical problems and promote research on enabling LLMs to simplify their reasoning via visual diagrams.
\end{abstract}


\section{Introduction}
\label{sec:intro}

\noindent\textbf{Formal language is central to modern mathematics, but diagrams and
constructive methods remain useful for developing and communicating mathematical
ideas.}

\begin{figure}[H]
  \centering
  \parbox{0.96\linewidth}{\centering\small\bfseries
    prove: every simple closed curve divides the plane into two regions for which it is the common boundary}
  \vspace{4pt}
  \includegraphics[width=0.7\linewidth]{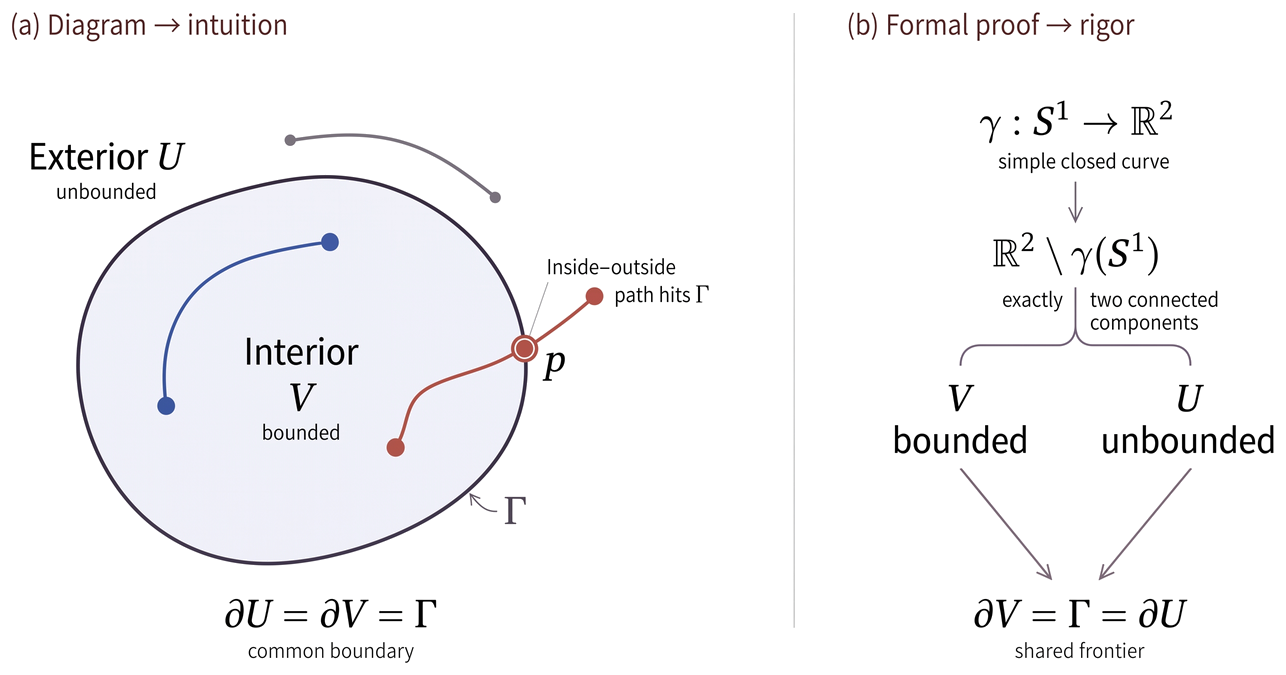}
  \caption{Visualizing the Jordan Curve Theorem. While the formal topological
  proof of this theorem is complex, a precise diagram delineating the inner and outer regions can make the reasoning process more intuitive.}
  \label{fig:jordan}
\end{figure}

In advanced research, text-based symbols can struggle to convey complex structural information directly, such as high-dimensional topological relations, morphism compositions (e.g., commutative diagrams), or manifold coordinate transformations. Precise mathematical figures can help compress these abstract relations, reduce symbolic ambiguity, and clarify proof strategies. Producing dependable academic figures, however, is demanding. It requires creators to understand mathematical abstractions and to control code construction precisely, since even small topological or metric deviations can derail the deductive process.

As illustrated in \Cref{fig:jordan}, the Jordan Curve Theorem provides a compact example of how formal statements relate to visual intuition.

This need for ``math-to-figure conversion'' exposes a weakness of current Large Language Models (LLMs). Requiring large models to generate precise mathematical figures (such as TikZ/SVG code) acts as a stress test, challenging four core capabilities:

\begin{enumerate}
  \item \textbf{Cross-modal spatial alignment.} Can the model correctly
        translate discrete, abstract mathematical symbols into continuous,
        concrete spatial topologies and coordinate geometries?
  \item \textbf{Global implicit geometric reasoning.} Without explicit
        coordinate prompts, can it autonomously solve implicit geometric
        constraints (e.g., relative positions of points, lines, and surfaces,
        tangency, orthogonality, or projection) and avoid spatial conflicts?
  \item \textbf{High-precision structured code generation.} Can it generate
        syntactically complete, logically self-consistent, and directly
        compilable vector graphics programs?
  \item \textbf{Semantic-level diagram verification.} In long-chain
        constructions, can it maintain strict consistency between local diagram
        elements and global mathematical theorems?
\end{enumerate}

\begin{figure}[H]
  \centering
  \includegraphics[width=\linewidth]{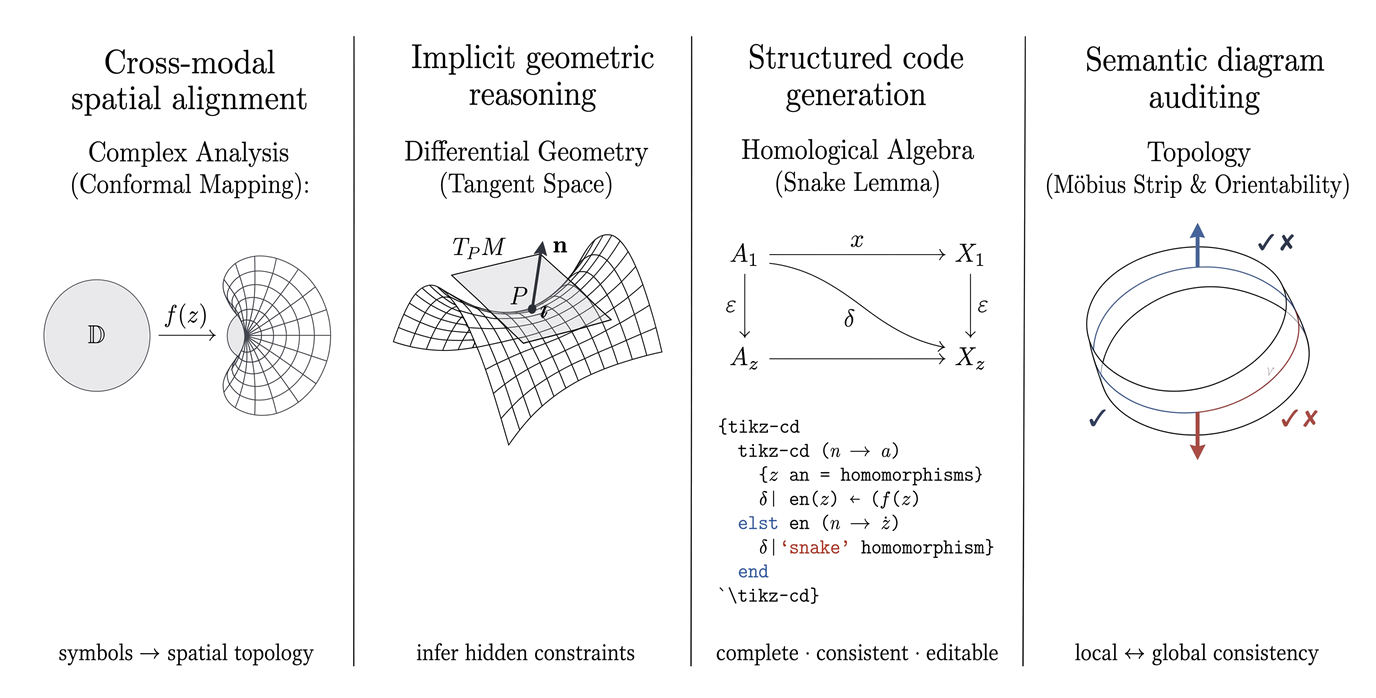}
  \caption{The four core capabilities required for mathematical figure generation.}
  \label{fig:capabilities}
\end{figure}

Most existing mathematical evaluations focus on scalar calculations or elementary geometry. As a result, they offer limited evidence about spatial construction in advanced mathematics and do not readily separate code errors from errors in geometric interpretation.

A small number of benchmarks study visual recognition and generation, but most focus on school-level material such as Euclidean geometry and elementary functions. MathVis instead covers 27 mathematical sub-disciplines, including differential geometry and complex analysis. Here, ``advanced'' refers to tasks whose mathematical objects or relations normally appear in upper-level undergraduate, graduate, or research-facing material and whose figures require more than elementary planar construction. Executable code and a Gold Intermediate Representation (Gold IR) help separate programming errors from errors in mathematical interpretation. On this benchmark, the evaluated LLMs still have difficulty with spatial and geometric construction. MathVis is intended as a common test bed for studying how models connect symbolic and visual representations in advanced mathematics.

Does AI-generated visualization help LLMs reason more effectively? The present evaluation does not establish a causal improvement in reasoning from access to diagrams. Instead, the results show that models often continue to rely on formal language and symbols even when a task calls for visual construction, and that a generated figure may contain the requested elements without organizing them into a representation that supports reliable reasoning. One plausible bottleneck is the limited amount of systematically paired training material that links advanced mathematical statements, executable diagrams, and diagram-assisted reasoning traces. MathVis does not audit model training corpora, so this explanation remains a research hypothesis rather than a measured conclusion. The benchmark provides tasks and evaluation signals for testing whether richer visual supervision can help future models use diagrams as part of the reasoning process rather than merely as final illustrations.

\section{Related Work}
\label{sec:related}

\noindent\textbf{Mathematical Figure Generation for Basic Education.} MathemaTikZ is one of the educational benchmarks closest to this work.
Originating from the Illustrative Mathematics curriculum, it curates 3,793
mathematical figures along with their natural language descriptions, problem
contexts, and TikZ implementations. It also provides systematic baseline
evaluations using GPT-4o, Claude 3.5 Sonnet, and Gemini 2.0
\citep{malik2025mathematikz}. The dataset covers K--12 graphics such as number
lines, bar charts, coordinate planes, and fraction strips. However, it focuses
on the editable reconstruction of curriculum content and does not involve proof
diagrams from advanced mathematical sub-disciplines. Another related work,
From Text to Visuals, translates text prompts into SVG mathematical
figures to generate problem-solving hints, using visual question answering to
check the semantic and syntactic quality of the figures
\citep{lee2025textvisuals}. Similarly, Math2Visual defines a specific
visual language and design space for math word problems, constructing 1,903
annotated visual examples to evaluate whether text-to-image models can preserve
pedagogical relationships \citep{wang2025math2visual}. Together, these studies
show how generated figures can support educational tasks, although they focus
mainly on hint diagrams and illustrations for basic mathematics. MathVis also
includes educational applications but applies the same verification protocol
to mathematical structures drawn from research-level material.

\noindent\textbf{Formal Geometry and Constraint Solving.} MagicGeo parses natural language into geometric constructions and generates
figures using geometric constraints; its MagicGeoBench contains 220 planar
geometry descriptions \citep{wang2025magicgeo}. Similarly, GeoLoom and its
accompanying dataset GeoNF automatically formalize natural language into
GeoLingua. They then use a coordinate solver to satisfy length, angle,
and construction dependencies, evaluating the generated results with structural
deviation metrics \citep{wei2025geoloom}. This approach offers a practical division of labor: language models parse the description, while formal languages and constraint solvers enforce geometric feasibility. However, the research scope
of these works remains focused on elementary planar geometry. This makes it
difficult to directly express advanced concepts, such as branches and winding
numbers in complex analysis, curvature and coordinate transformations in
differential geometry, or genus and homotopy relations in topology. MathVis
borrows the idea of using intermediate representations and constraint
verification. However, it extends the scope of these constraints to a
cross-disciplinary Gold IR, evaluating them through question-level hard and soft
atomic predicates.

\noindent\textbf{Theorem Explanation and Scientific Figure Program Synthesis.} TheoremExplainAgent takes theorems as input and uses a planning agent alongside a
coding agent to generate long-form theorem explanation videos based on Manim. It
evaluates both factual and visual quality using 240 theorems from
TheoremExplainBench and five types of automatic metrics
\citep{ku2025theoremexplain}. Like MathVis, it emphasizes theorem structures and
verifiable visual expressions. However, its outputs do not specifically focus on static figures for advanced fields such as complex analysis, differential geometry, and topology.

In the realm of static graphics, AutomaTikZ learns text-to-TikZ scientific
vector graphic synthesis from paper figures and their surrounding contexts,
releasing approximately 120,000 captioned TikZ figure pairs
\citep{belouadi2023automatikz}. DeTikZify focuses on recovering TikZ programs
from scientific figures or sketches \citep{belouadi2024detikzify}, while TikZero
uses image representations to bridge unaligned graphic programs and image
caption data, achieving zero-shot text-guided graphic program synthesis
\citep{belouadi2025tikzero}. These works provide data-construction pipelines based on figures and surrounding text from papers and arXiv preprints.

Several recent works are also directly relevant to the present scope. Visual Sketchpad studies sketching as a visual chain of thought for multimodal reasoning \citep{hu2024visualsketchpad}. SGP-Bench asks whether language models can understand symbolic graphics programs without relying on vision encoders \citep{qiu2025sgpbench}. VGBench evaluates both understanding and generation of vector graphics across formats such as SVG and TikZ \citep{zou2024vgbench}. On the visual-mathematics side, MathVerse and MATH-Vision study whether multimodal systems truly use diagrams in mathematical reasoning \citep{zhang2024mathverse,wang2024mathvision}. These works are complementary to MathVis: they study visual mathematical reasoning or general vector-graphics competence, whereas MathVis focuses on generating executable, verifiable figures from advanced mathematical statements. To avoid name confusion, we distinguish MATH-Vision \citep{wang2024mathvision} from the later diagram-generation benchmark Math-Vision Diagrams \citep{kashyap2026mathvisiondiagrams}.

\begin{figure}[H]
  \centering
  \makebox[\linewidth][c]{\includegraphics[width=1.13\linewidth]{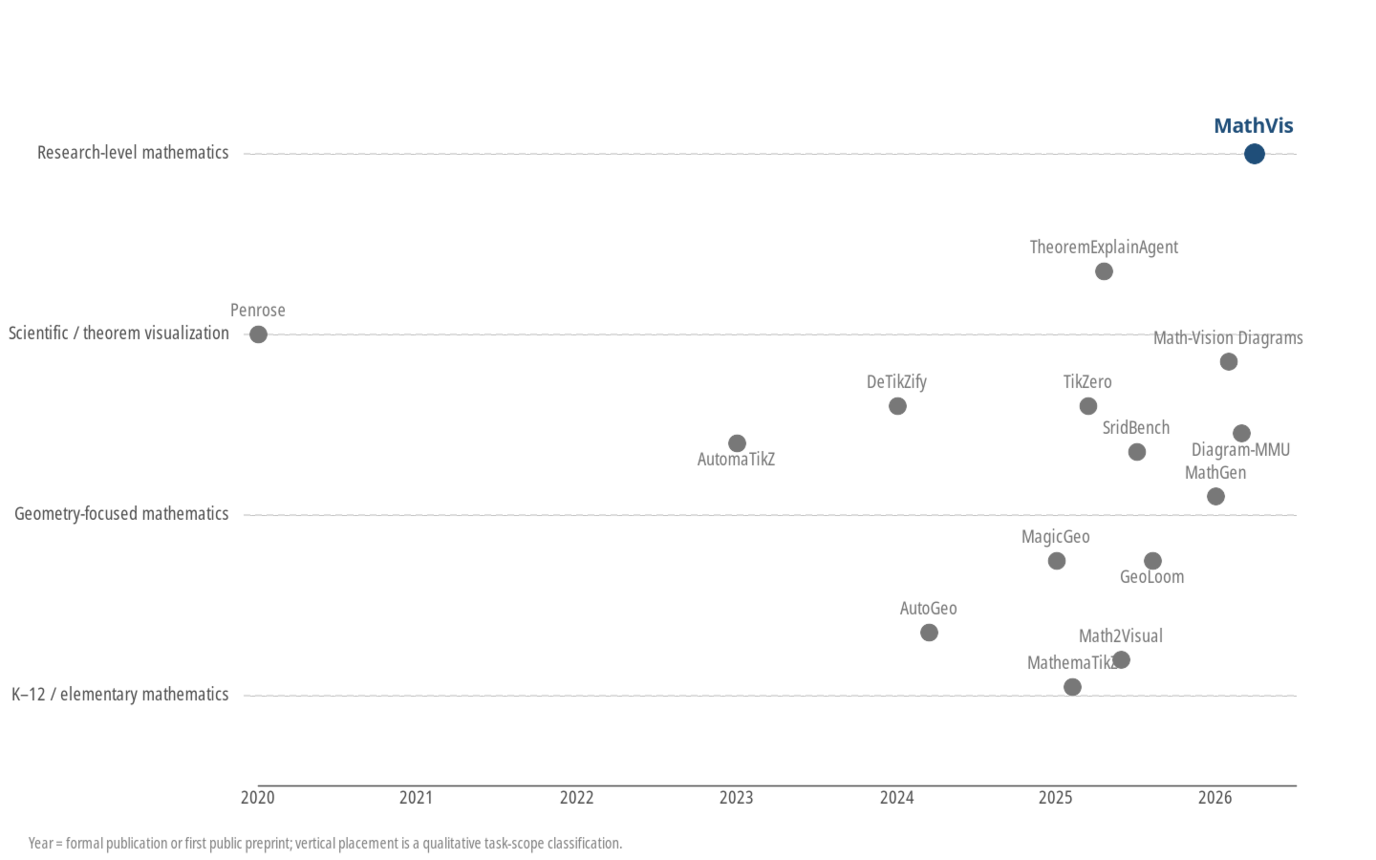}}
  \caption{Representative benchmarks positioned by first public-release date and qualitative research scope.}
  \label{fig:related-scope}
\end{figure}

Relative to the target setting of MathVis, existing resources leave two practical gaps; this comparison concerns task coverage rather than the overall value of those resources.

\begin{enumerate}
  \item \textbf{Educational level.} Research on mathematical figure generation
        has largely focused on K--12 and middle school geometry (e.g.,
        AutoGeo, MagicGeo, GeoLoom, MathemaTikZ, Math2Visual)
        \citep{huang2024autogeo,wang2025magicgeo,wei2025geoloom,malik2025mathematikz,wang2025math2visual}.
  \item \textbf{Disciplinary depth.} Works that do cover ``theorems'' or ``advanced scientific figures'' (e.g., TheoremExplainBench, the AutomaTikZ series, SridBench, Penrose) may include non-mathematical material, may not be organized by mathematical branch, or may offer limited coverage of advanced fields such as complex analysis, differential geometry, and topology \citep{ku2025theoremexplain,belouadi2023automatikz,chang2025sridbench,ye2020penrose}.
\end{enumerate}

\section{Method}
\label{sec:method}

\setcounter{subsection}{1}

\subsection{Benchmark Construction}

\begin{figure}[H]
  \centering
  \includegraphics[width=\linewidth]{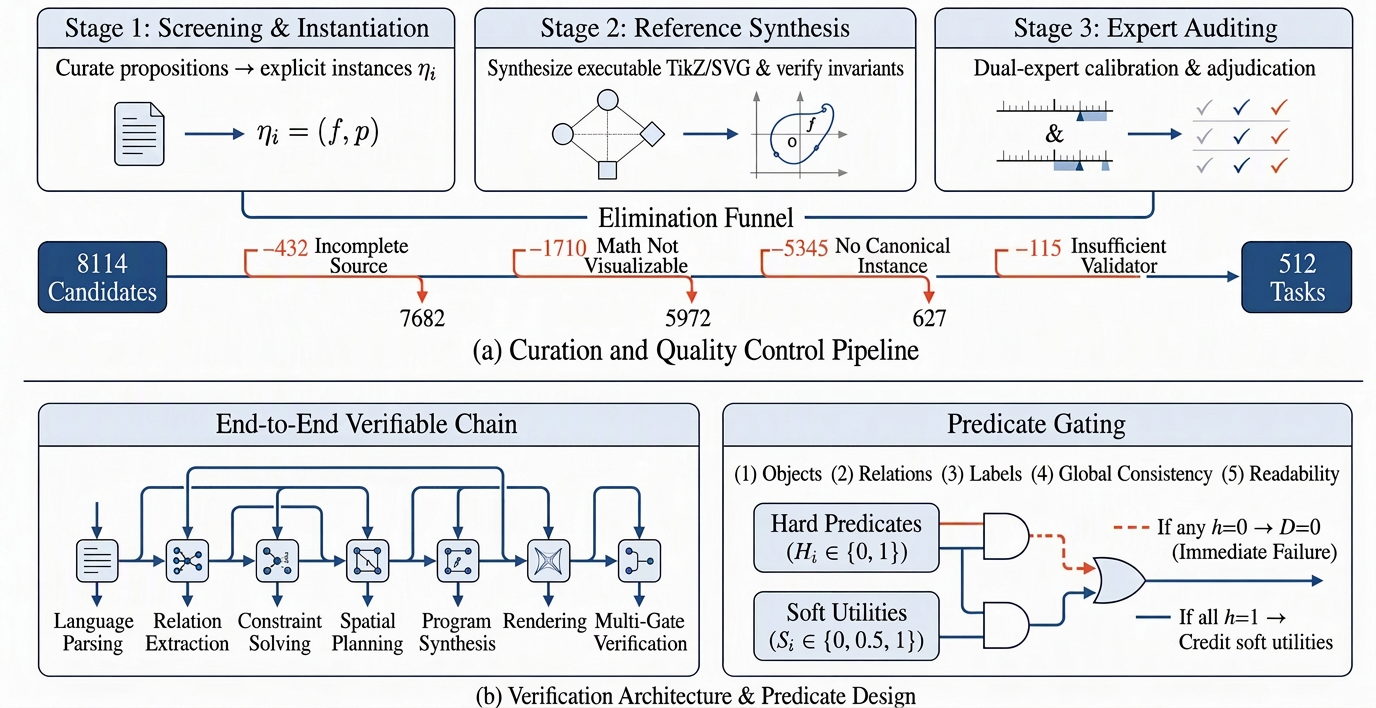}
  \caption{Curation pipeline and verification architecture.}
  \label{fig:curation}
\end{figure}

\Cref{fig:curation} summarizes the three-stage curation pipeline and the verification architecture that together produce the frozen task packages.

\subsection{Evaluation Set Coverage}

\begin{figure}[htbp]
  \centering
  \includegraphics[width=\linewidth]{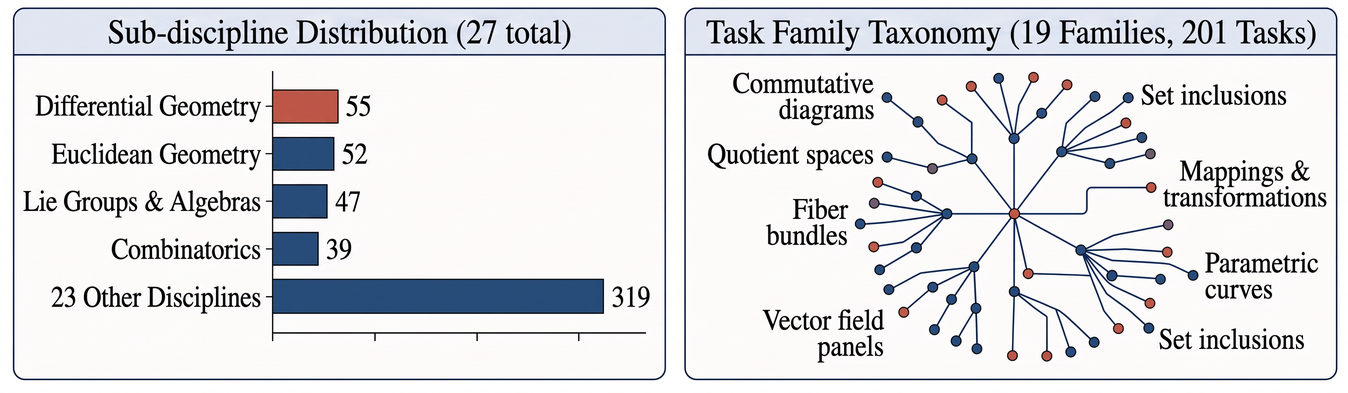}
  \caption{Dataset composition.}
  \label{fig:composition}
\end{figure}

The benchmark contains 512 questions from 14 source collections and 27
mathematical sub-disciplines; 201 of these currently have task-family metadata,
covering 19 families in total, including local geometric relations, proof
auxiliary constructions, mappings and coordinate transformations, parametric
curves and limits, set inclusion and coverings, commutative diagrams, quotient
spaces, fiber bundles, vector field panels, and counterexample constructions.
Before freezing, the questions underwent source tracking, sub-discipline
coverage, task-package completeness, and automatic/expert auditing to form a
frozen evaluation-only set with hidden evaluator assets. In this paper, the task-family labels are
descriptive metadata only; they are not used to define train/dev/test splits,
because the current benchmark release does not define training or development splits. This
evaluation set covers a variety of mathematical structures but is not claimed to
represent the overall difficulty distribution. The sub-discipline composition is
summarized in \Cref{fig:composition}.

\subsection{Task Packages and Annotation Quality}

We establish a ground-truth evaluation framework, as detailed in \Cref{app:metrics}.

Task-package and annotation quality is managed through a division of expert
responsibilities. Task authors select mathematical objects, define the target
capabilities, write the problem statements, and construct the figures. Domain
reviewers then examine the geometric relations, the correspondence between the
mathematics and the figure, question difficulty, and solution uniqueness. Formal
verification experts convert accepted solutions into Gold IR, atomic predicates,
and automated checks, while adjudicators resolve disagreements and edge cases
before the final version is frozen.

Quality control spans candidate generation from traceable sources, independent
problem solving, assessment of whether a figure is necessary, Gold IR
construction, predicate and validator testing, independent dual-expert review,
adjudication, model trial runs, and final freezing. Eliminated questions and the
reasons for revision are retained. Questions may be removed because of ambiguous
prompts, unnecessary figures, non-unique answers, incomplete proofs, insufficient
validation evidence, or mismatched difficulty; recording these decisions makes
the review process auditable.

\subsection{Evaluation Protocol}
\label{sec:protocol}

The repeated-generation metrics \passatthree{} and \passthree{} describe two possible deployment policies. \passatthree{} counts a question as successful when at least one of its three outputs strictly passes and is relevant when a validator can select among candidates. \passthree{} counts a question as successful only when all three outputs strictly pass and measures repeated-generation stability. Their definitions are retained in \Cref{app:metrics}.

Rubric coverage is defined using the question-specific number of atomic rubrics $M_i$ and their weights $w_{ij}$, rather than assuming a fixed number of rubrics for every task. The normalized score $R_{i,m,r}$ preserves partial diagnostic information, whereas strict delivery requires the technical gate, every mathematical hard predicate, and complete soft-constraint satisfaction. Aggregate rubric coverage is not tabulated in this version because it is not included in the retained workbook.

\section{Experiments}
\label{sec:experiments}

\subsection{Models and Runs}

This paper defines a Harness as a fixed execution protocol that wraps model calls.
A Harness simultaneously specifies prompt templates, models and providers, sampling parameters, output formats, code extraction methods, compilers and renderers, validators, timeout limits, retry strategies, and run log fields. The evaluation includes the 16 model configurations listed in
\Cref{tab:family-scorecard}; no other model result is included in the scorecard or ranking figure. All models receive the same task prompts and output contracts, without reference figures provided.

\begin{sotanote}[Task prompt]
  Do not search the internet or call other existing drawing tools; rely solely
  on the model's capabilities to complete this question. Generate a mathematical
  diagram based on the following mathematical description, ensuring mathematical
  accuracy; output only the figure; do not add objects conflicting with the
  problem meaning; all specified labels must be clearly legible. The result must
  be delivered as a code file, choose one of two: (1)~a complete TikZ standalone
  document that can be directly saved as \texttt{.tex} and compiled; (2)~a
  complete SVG document that can be directly saved as \texttt{.svg} and
  rendered.
\end{sotanote}

\subsection{Main Results}

\begin{figure}[htbp]
  \centering
  \includegraphics[width=0.88\linewidth]{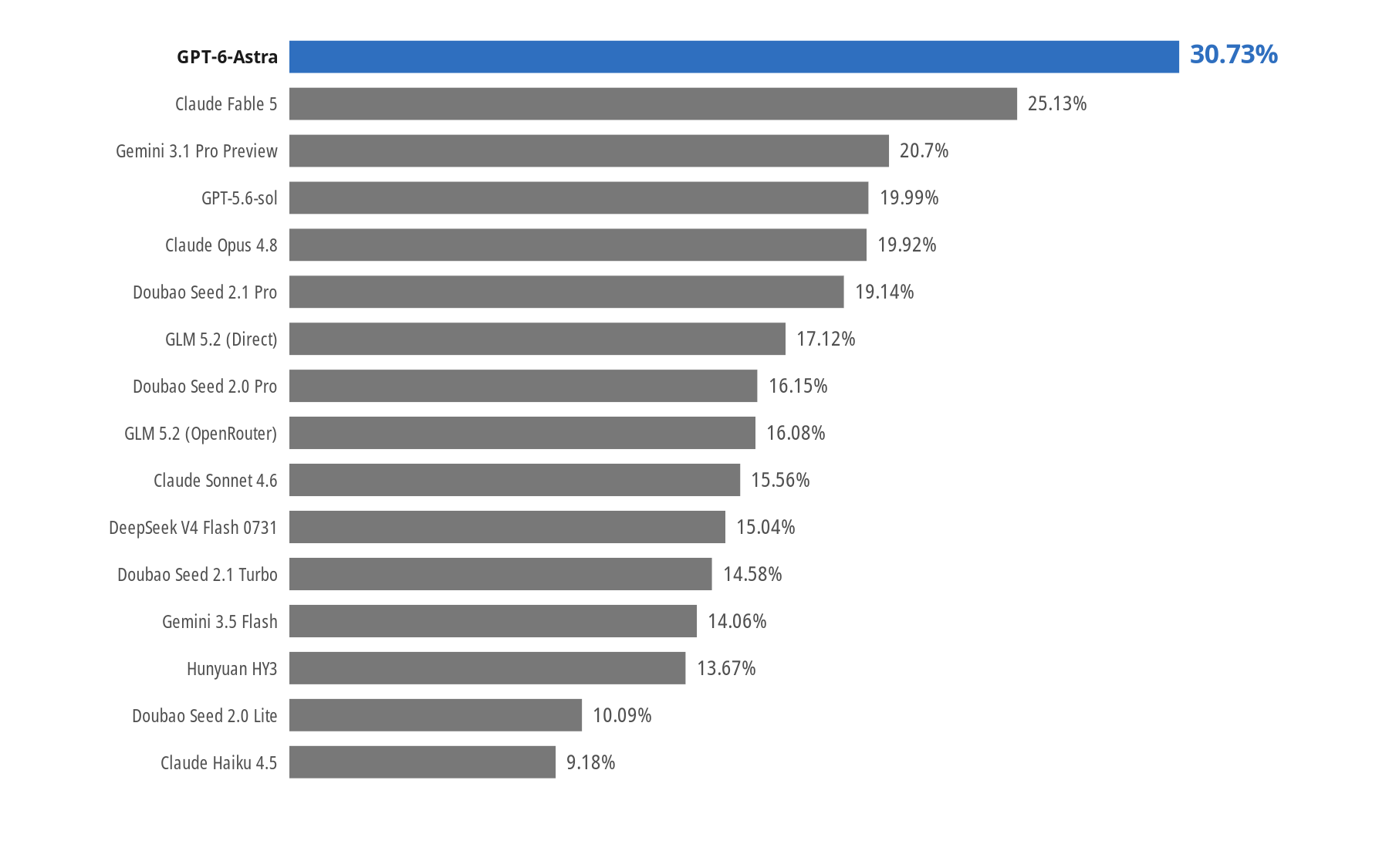}
  \caption{Observed strict $D$-PSR rates for the 16 evaluated model configurations, ordered from highest to lowest. Only models listed in the accompanying evaluation workbook are shown.}
  \label{fig:family-scorecard}
\end{figure}

\Cref{fig:family-scorecard} visualizes the observed strict deliverability rates, and \Cref{tab:family-scorecard} reports six metrics from the retained evaluation data: \passone{}, $D$-PSR, the technical gate rate $T$, $S$-PSR, \passatthree{}, and \passthree{}. Comparisons are made under the same 512 tasks, prompts, renderers, and three-rollout protocol.

The top three configurations by \passone{} are summarized below for quick reference in the main text.

\begin{table}[H]
  \centering
  \caption*{Top three configurations extracted from Table~\ref{tab:family-scorecard}.}
  \footnotesize
  \setlength{\tabcolsep}{4.5pt}
  \begin{tabular}{@{}lcccccc@{}}
    \toprule
    \sotaheader{Model} &
    \sotaheader{\passone} &
    \sotaheader{$D$-PSR} &
    \sotaheader{$T$} &
    \sotaheader{$S$-PSR} &
    \sotaheader{\passatthree} &
    \sotaheader{\passthree} \\
    \midrule
    GPT-6-Astra & 31.25 & 30.73 & 94.40 & 31.20 & 38.28 & 24.22 \\
    Claude Fable 5 & 26.56 & 25.13 & 84.11 & 26.07 & 29.30 & 19.14 \\
    Gemini 3.1 Pro Preview & 22.85 & 20.70 & 78.13 & 20.70 & 27.34 & 15.23 \\
    \bottomrule
  \end{tabular}
\end{table}


The scorecard should not be read as establishing a stable global model ranking.
The comparison is limited to the frozen prompts, renderers, and run settings used in this evaluation. By construction, $S$-PSR is no lower than $D$-PSR because every strict delivery has $S=1$, whereas partial soft credit can also be retained when $D=0$. Across the 16 evaluated models, the observed $D$-PSR ranges from $9.18\%$ to $30.73\%$; GPT-6-Astra has the highest observed rate, followed by Claude Fable 5 at $25.13\%$. These are descriptive results for this evaluation set, not a universal ranking.

\setcounter{subsection}{5}

\subsection{Harness and Agent Challenge Examples}

The data in \Cref{tab:family-scorecard} come from non-agent baselines: single-round code generation with zero tool calls, aiming to test the model's own diagram-generation capability under the benchmark constraints. To examine a more tool-augmented usage setting, this paper treats agent and direct baseline as two independent experimental configurations, fixing a set of challenge examples for subsequent review under the same budget. Only if the agent's final artifact still satisfies $D = 0$ after the pre-registered three-round budget is it recorded as ``agent still fails'', retaining all intermediate drafts for auditing.

\begin{table}[htbp]
  \caption{Agent challenge examples. Direct baselines and independently
    evaluated agent runs are recorded separately under the same tasks, prompts,
    and budget definitions.}
  \label{tab:agent-cases}
  \small
  \begin{tabularx}{\linewidth}{@{}lYY@{}}
    \toprule
    \sotaheader{Question} & \sotaheader{Mathematical bottleneck} &
    \sotaheader{Direct baseline evidence} \\
    \midrule
    DG-FIG-01 &
      Coordinate image arrows, transformation directions, and inverse mapping labels &
      Archived outputs reverse or omit required mappings; displayed artifacts fail the strict delivery gate \\
    DG-FIG-02 &
      Four open semi-circular domains, quadrant intersections, and radical signs &
      Archived outputs include rendering failures and incomplete domain constructions \\
    CV2-FIG-13-01 &
      Disk interior angles, symmetry axes, and the $2\alpha$ relation &
      Archived outputs include missing angle relations and rendering failures \\
    \bottomrule
  \end{tabularx}
\end{table}

These cases target errors that may not be resolved solely by renderer repairs: syntactically valid figures may still reverse mappings, place intersections in the wrong quadrants, or miss geometric constraints.

\subsection{Detailed Case Analysis}

To make the evidence directly visible in the main text, we include two detailed agent test instances below: an application of Rouch\'e's Theorem and a Mandelbrot set construction. Together, these two cases suggest that current agent settings do not yet yield stable gains over the direct baselines. In the Rouch\'e case, even a tool-augmented run still misses hard mathematical constraints and does not produce a reliable strict pass; in the Mandelbrot case, the archived outputs remain vulnerable to truncation, incomplete external-ray structure, and missing co-landing relations. The common pattern is that extra steps or tooling do not reliably repair the core mathematical semantics once the construction itself is already off track.

\subsection{Agent Test Instance: Application of Rouch\'e's Theorem}
\label{sec:rouche}

Rouché's Theorem is a standard result in complex analysis for determining the number of zeros of an analytic function within a region; it also follows from the Argument Principle. The theorem states that if an analytic function is perturbed while the dominant term remains larger on the boundary, then the total number of zeros is unchanged.

Let $C$ be a simple closed curve, and let functions $f(z)$ and $g(z)$ satisfy
the following conditions:

\begin{enumerate}
  \item analytic inside and on $C$ (or $f$ is meromorphic inside and continuous
        to the boundary);
  \item satisfy $|f(z)| > |g(z)|$ on the boundary $C$.
\end{enumerate}

Then the functions $f(z)$ and $f(z) + g(z)$ have the same number of zeros (counting multiplicities) inside $C$, that is, $N(f + g, C) = N(f, C)$.

This theorem has an intuitive geometric interpretation, informally known as the "Dog-Walking Theorem":

\begin{itemize}
  \item Imagine a tree at the origin, and a person (representing $f(z)$) walking
        a dog (representing $f(z) + g(z)$) around the tree.
  \item The length of the leash represents $|g(z)|$, and the distance from the
        person to the tree represents $|f(z)|$.
  \item If the leash length is always less than the distance from the person to
        the tree ($|g| < |f|$), then the number of times the dog circles the
        tree is the same as the number of times the person circles the tree.
  \item This means their winding numbers around the origin are identical, thus
        the number of zeros is the same.
\end{itemize}

\begin{sotanote}[Prompt]
  Generate an executable figure to illustrate the application of Rouch\'e's
  Theorem for $f(z) = z^3$ and $g(z) = z + 1$ on $C = \{z : |z| = 2\}$.
\end{sotanote}

\begin{figure}[htbp]
  \centering
  \includegraphics[width=0.62\linewidth]{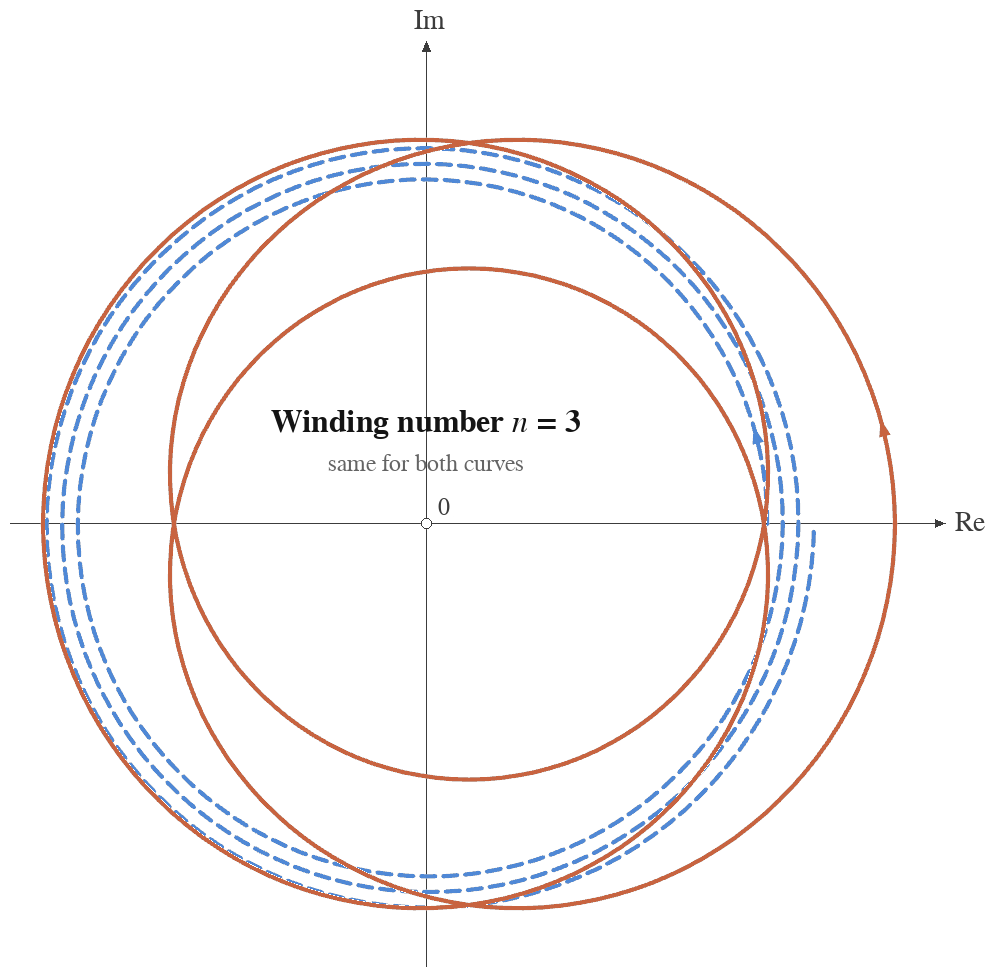}
  \caption{Expert reference figure for the Rouch\'e's Theorem task: the images
    of $f(z) = z^3$ and $f(z) + g(z) = z^3 + z + 1$ on $C = \{z : |z| = 2\}$
    wind three times around the origin, so both have three zeros inside $C$.}
  \label{fig:rouche-reference}
\end{figure}

The expert reference construction is shown in \Cref{fig:rouche-reference}.

\begin{table}[htbp]
  \caption{Atomic constraint card for the Rouch\'e's Theorem task. Hard
    constraints gate delivery; soft-constraint scores are credited only once
    every hard constraint passes.}
  \label{tab:rouche-rubric}
  \footnotesize
  \begin{tabularx}{\linewidth}{@{}llYYl@{}}
    \toprule
    \sotaheader{ID} & \sotaheader{Type} & \sotaheader{Atomic decision question} &
    \sotaheader{Evidence and allowed equivalent forms} & \sotaheader{Weight / code} \\
    \midrule
    CA-H1 & Hard &
      Is $C$ correctly specified as the circle centered at $0$ with radius $2$? &
      A labeled equation or domain-plane circle, allowing global translation and
      uniform scaling when geometry is shown &
      \texttt{DOMAIN} \\
    CA-H2 & Hard &
      Are the images of $C$ under $f(z)=z^3$ and
      $f(z)+g(z)=z^3+z+1$ correctly drawn and labeled? &
      Curve geometry and function--curve binding, allowing equivalent color schemes
      and line styles &
      \texttt{FUNCTION} \\
    CA-H3 & Hard &
      Does the figure or its associated formula give the correct boundary comparison,
      $|z + 1| \leq 3 < 8 = |z^3|$ for $z \in C$? &
      Boundary evaluation on $|z| = 2$, accepting equivalent algebraic
      presentations of the same inequality chain &
      \texttt{BOUNDARY} \\
    CA-H4 & Hard &
      Does the figure indicate that $f(z)+g(z)$ has exactly three zeros inside $C$,
      either by marking the three zeros or by showing a winding number of three? &
      Count and region membership of zero markers, or the winding number of the image
      curve about the origin; exact root coordinates are not required &
      \texttt{ZEROCOUNT} \\
    CA-H5 & Hard &
      Regardless of the graphical perspective, does the conclusion state that $f$
      and $f + g$ have the same number of zeros inside $C$, both being three
      counting multiplicities? &
      Text or formula extraction, accepting equivalent phrasing &
      \texttt{CONCLUSION} \\
    \midrule
    CA-S1 & Soft &
      Are the chosen contours, regions, winding cues, and any zero markers easy to distinguish? &
      Scored as $0$, $0.5$, or $1$ by experts based on public anchors &
      0.30 \\
    CA-S2 & Soft &
      Are coordinate axes, function names, contour labels, and formulas clear,
      with no incorrect spatial bindings? &
      Text recognition and overlap check, confirmed by experts &
      0.25 \\
    CA-S3 & Soft &
      Do inequalities form clear visual associations with contours, rather than
      pointing to irrelevant regions? &
      Spatial binding or leader line evidence &
      0.25 \\
    CA-S4 & Soft &
      Does the layout avoid non-critical intersections, clipping, and label
      collisions? &
      Bounding box detection and expert readability check &
      0.20 \\
    \bottomrule
  \end{tabularx}
\end{table}

For each of CA-S1 to CA-S4, the atomic score $s_j=1$ means that the condition is fully satisfied, $s_j=0.5$ means that the intended mathematical association remains recoverable but requires extra effort to discern, and $s_j=0$ means that the association is missing or misleading. Their weighted aggregate is the task-level soft score $S\in[0,1]$. The strict delivery indicator uses the same definition as the rest of the benchmark:

\begin{equation}
  D=T\,H_{\mathrm{math}}\,\mathbb{1}[S=1].
  \label{eq:hardgate}
\end{equation}

Thus, if CA-H3 fails, $H_{\mathrm{math}}=0$ and $D=0$ regardless of $S$; technical validity and complete presentation-constraint satisfaction are also required for strict delivery. The word ``soft'' refers to the graded diagnostic scale, not to optionality in the full-specification delivery gate: partial credit remains diagnostic and does not create a strict pass.

This rubric rewards clarity only once the theorem-related mathematical content is viable.

\begin{figure}[p]
  \centering
  \begin{minipage}{0.92\linewidth}
    \centering
    \includegraphics[width=\linewidth]{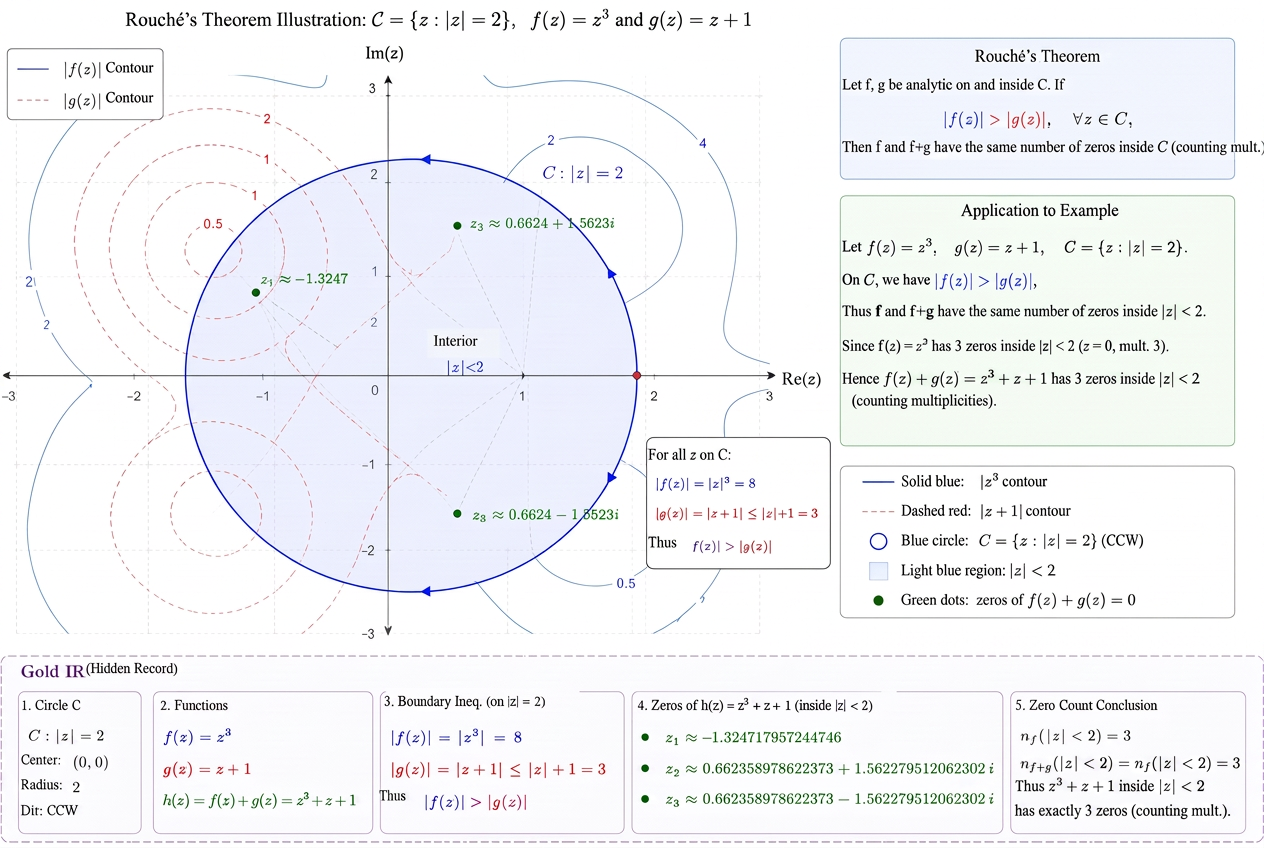}
    \captionof{figure}{GPT-5.6-sol error diagram for the Rouch\'e's Theorem task. All labels in the archived artifact are presented in English.}
    \label{fig:rouche-gpt}
  \end{minipage}

  \vspace{10pt}

  \begin{minipage}{0.92\linewidth}
    \centering
    \includegraphics[width=\linewidth]{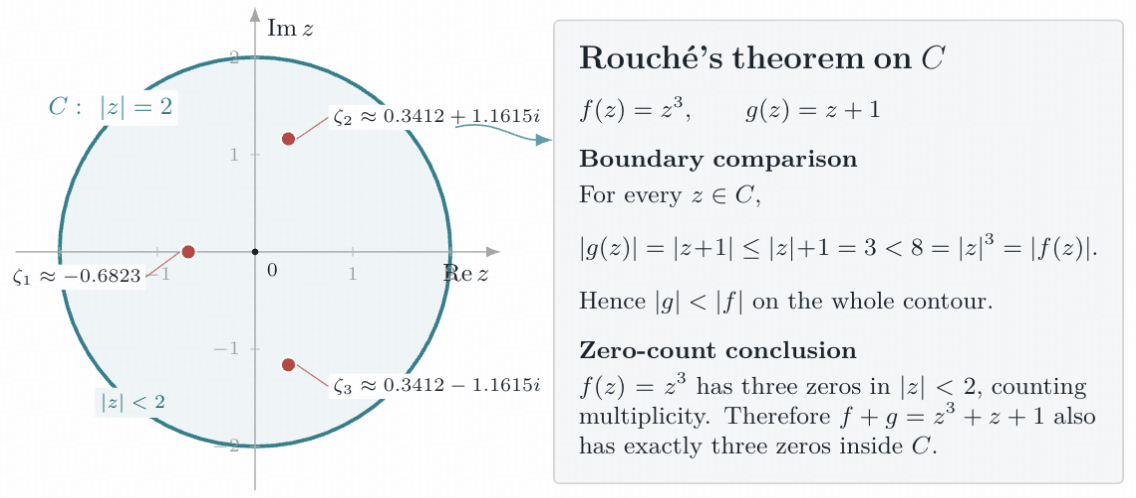}
    \captionof{figure}{Error diagram for the Rouch\'e's Theorem task generated by Claude Fable 5 through the Claude Code agent environment. Claude Fable 5 is the underlying model; Claude Code identifies the agent Harness.}
    \label{fig:rouche-claude}
  \end{minipage}
\end{figure}

As shown in \Cref{fig:rouche-gpt}, the archived GPT-5.6-sol output contains the following errors.

\paragraph{GPT-5.6-sol deductions.}

\begin{enumerate}
  \item $C$ is not a simple circle centered at $0$ with radius $2$, violating
        CA-H1, so the score is $0$.
  \item The rendered curves do not correctly depict and label both images
        $f(C)$ and $(f+g)(C)$, violating CA-H2, so the score is $0$.
  \item The figure neither marks the three zeros of $f(z)+g(z)$ correctly inside
        $C$ nor provides a valid winding-number representation with three turns,
        violating CA-H4, so the score is $0$.
\end{enumerate}

The output in \Cref{fig:rouche-claude} was generated by the Claude Fable 5 model through the Claude Code agent Harness and has a different set of errors.

\paragraph{Claude Fable 5 deductions.} The figure does not correctly draw and label both images $f(C)$ and $(f+g)(C)$, violating CA-H2, so the score is $0$. It also does not state, through the figure or an associated formula, that $f$ and $f + g$ have the same number of zeros inside $C$, namely three when counted with multiplicity, violating CA-H5, so the score is $0$.

\begin{keyfinding}[Summary]
  The agent + Harness configuration did not yield a consistent improvement.
  Step-by-step guidance that first asks the model to draw the images of
  $f(z) = z^3$ and $f(z) + g(z) = z^3 + z + 1$ on
  $C = \{z : |z| = 2\}$, then identify the loops and relate them to the zeros,
  appeared to improve some outputs. This observation is qualitative and should
  not be interpreted as a measured effect size.
\end{keyfinding}

\subsection{Agent Test Instance: Mandelbrot Set}
\label{app:mandelbrot}

\begin{sotanote}[Prompt]
  Draw equipotential lines and rational external rays outside the Mandelbrot set
  according to parametric B\"ottcher coordinates, and correctly express the
  period 1--4 branches and the co-landing points such as $1/7$, $2/7$, $3/15$,
  and $4/15$.
\end{sotanote}

\begin{figure}[p]
  \centering
  \begin{minipage}{0.60\linewidth}
    \centering
    \includegraphics[width=0.82\linewidth]{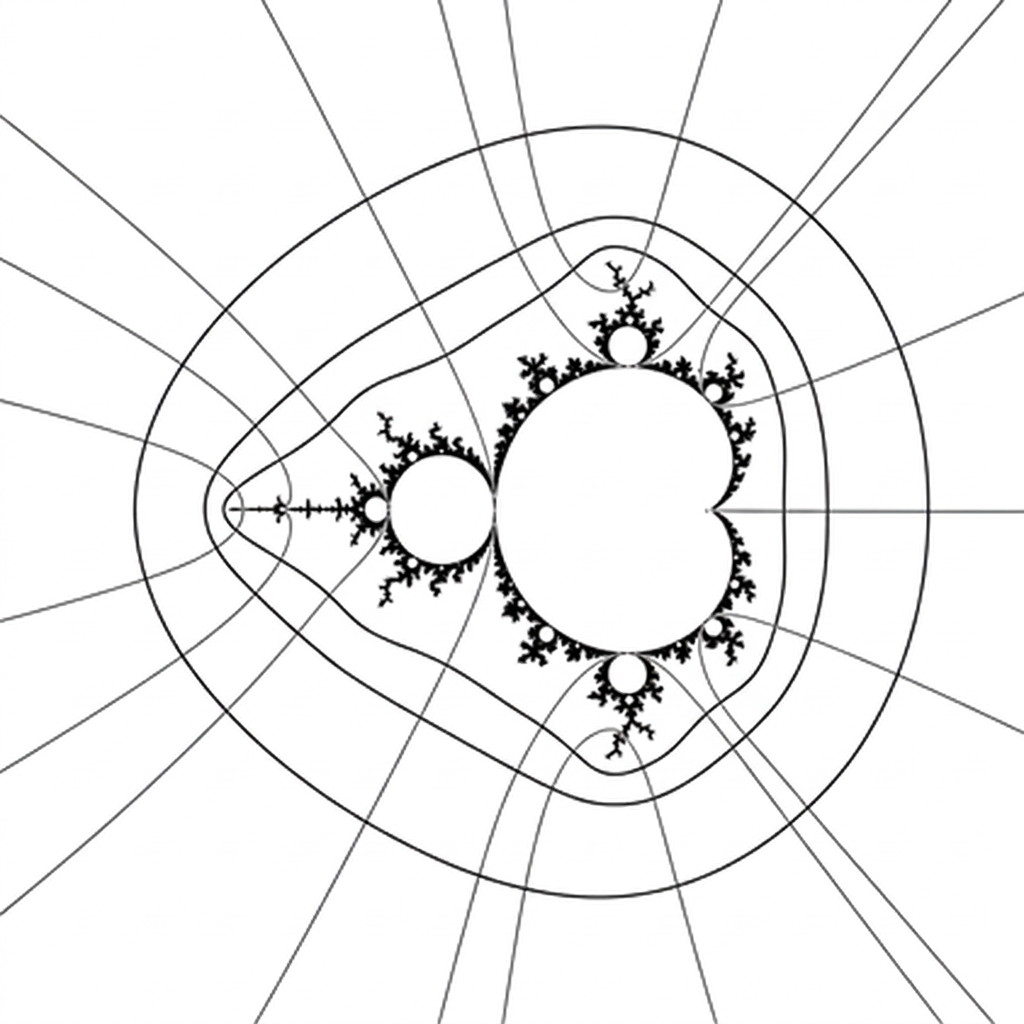}
    \captionof{figure}{Expert reference figure for the Mandelbrot set task: the set with several concentric equipotential lines and rational external rays arranged counterclockwise.}
    \label{fig:mandelbrot-reference}
  \end{minipage}

  \vspace{12pt}

  \begin{minipage}{0.98\linewidth}
    \centering
    \begin{minipage}[t]{0.48\linewidth}
      \centering
      \includegraphics[width=\linewidth]{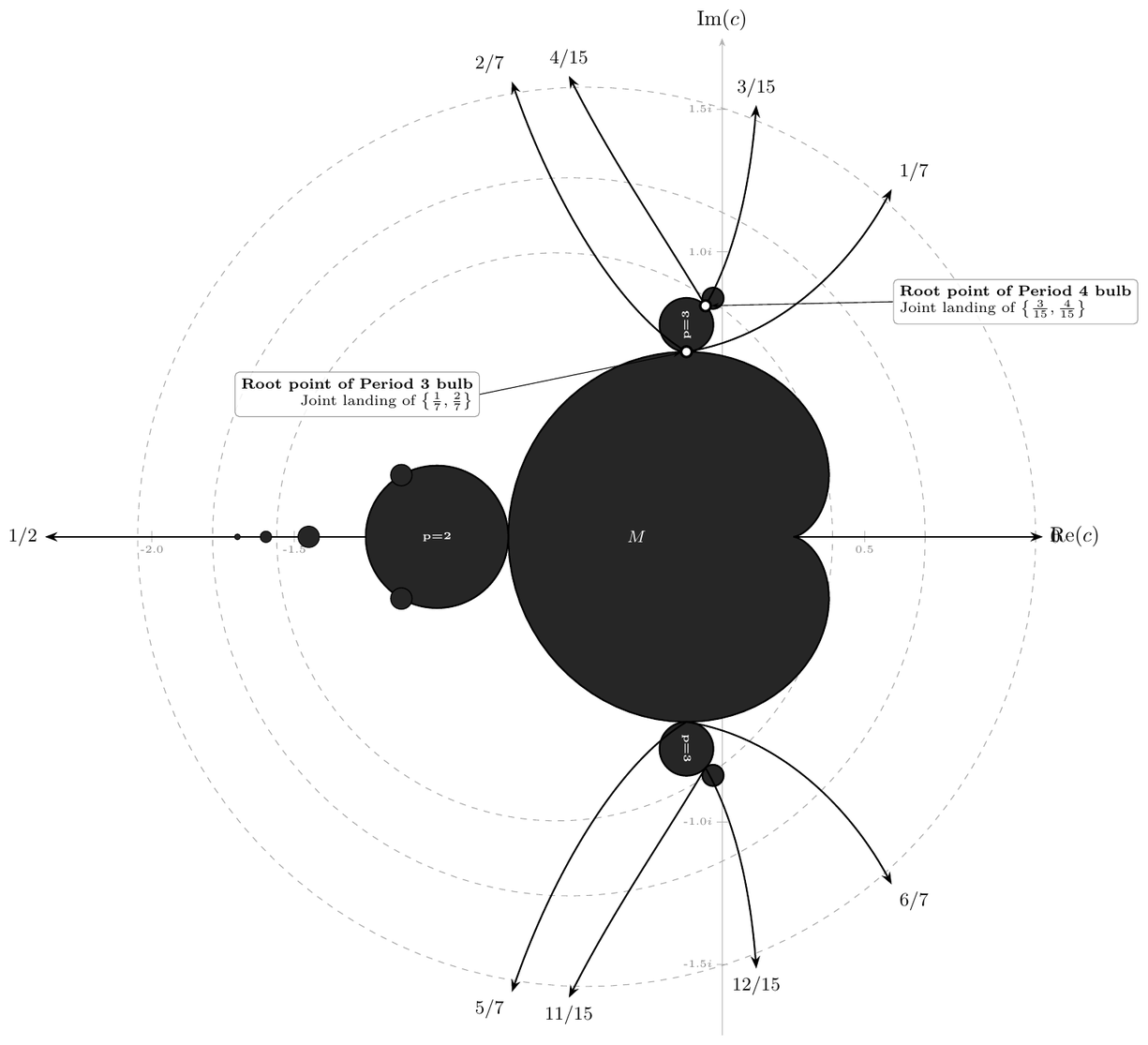}\\[-2pt]
      \textbf{(a)} Gemini 3.5 Flash
    \end{minipage}\hfill
    \begin{minipage}[t]{0.48\linewidth}
      \centering
      \includegraphics[width=\linewidth]{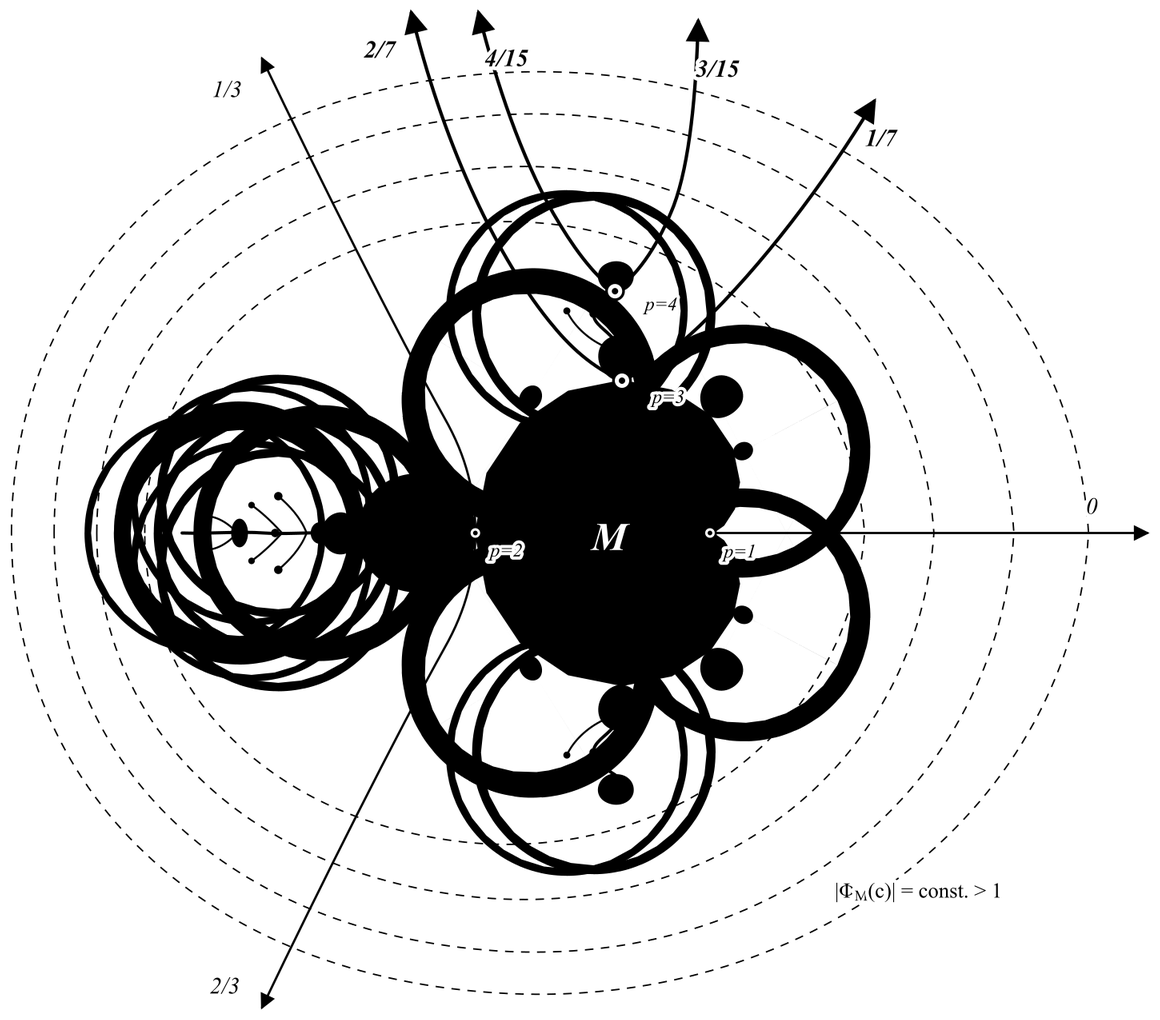}\\[-2pt]
      \textbf{(b)} GPT-5.6-sol
    \end{minipage}
    \captionof{figure}{Side-by-side comparison of model outputs for the Mandelbrot set task. Gemini 3.5 Flash (left) misses the required B\"ottcher equipotentials and misplaces the period-4 branch; GPT-5.6-sol (right) produces an inaccurate boundary and incomplete characteristic rays.}
    \label{fig:mandelbrot-model-comparison}
  \end{minipage}
\end{figure}

Mandelbrot set external rays are defined by parametric B\"ottcher coordinates;
rational angle rays with periods 1, 2, 3, and 4 land at specific root points of
$\partial M$ and encode hyperbolic components. The figure simultaneously presents
the Mandelbrot set, several concentric equipotential lines, and rational external
rays arranged counterclockwise; the co-landing point pairs like $1/7$, $2/7$,
$3/15$, $4/15$ and their period classifications must be correct. The graphical relations should support the stated conclusion rather than merely replicating the visual appearance.

The expert reference is shown in \Cref{fig:mandelbrot-reference}.

\paragraph{Gemini 3.5 Flash.} Insufficient iteration and boundary precision;
equipotential lines are not B\"ottcher equipotential lines; the period ray family
is incomplete, and the topological position of the period-4 branch is drawn
incorrectly.

\paragraph{GPT-5.6-sol.} Insufficient iteration precision; the Mandelbrot set is
drawn as a combination of a disk and a thick annulus, missing the true fractal
boundary; co-landing points are not strict enough; equipotential lines are not
calculated from parametric B\"ottcher coordinates, and only partial characteristic
rays of periods 3 and 4 are drawn, which cannot be called a complete
construction.

\Cref{fig:mandelbrot-model-comparison} compares the two archived model outputs side by side for direct inspection.

\paragraph{Claude Opus 4.8.} The generation process was truncated after exceeding
the 128,000-token budget, returning only thinking; the final code was empty,
triggering \texttt{TECH\_NO\_OUTPUT}.

\sotadivider

\section{Discussion}
\label{sec:discussion}

The results suggest potential implications for both training and deployment. For
training, the observed failure separation suggests that compilation-only objectives may be insufficient; future systems could use
supervision that jointly targets mathematical relations, label binding, global
constraint consistency, and readable spatial organization. Gold IR and atomic
predicates make these failure modes observable, enabling curriculum design,
hard-negative construction, verifier-guided fine-tuning, and targeted repair
rather than undifferentiated image-level optimization.

For deployment studies, the appropriate metric may depend on the operating mode. Systems
that can generate several candidates and run a trusted validator may benefit from
candidate selection, whereas systems that deliver a single artifact directly need
a strict output-level gate. In the retained results, $T$ diagnoses technical
executability, $S$-PSR retains partial soft information, and $D$-PSR measures strict
delivery. Soft scores can help prioritize repairs, but they should not override mathematical or technical
hard gates.

The gap between relatively high rendering success and lower strict
$D$-PSR motivates evaluating a layered deployment architecture in future work: compile and render
first, verify mathematical predicates second, and assess readability only after
hard correctness is established. The present experiments do not measure the latency, cost, or user-level benefit of such an architecture. Repair budgets, validator access, latency, and
cost should therefore be treated as explicit deployment parameters rather than
hidden implementation details. Direct and tool-augmented Harnesses should be
reported separately so that improvements can be attributed to model capability,
verification, or iterative repair.

\section{Limitations and Responsible Use}
\label{sec:limitations}

Current evidence supports transparent comparisons among the evaluated frozen
configurations, but it does not support unbiased population-level claims about
all mathematical problems, model families, or agent settings. The 512 tasks are
not a random sample of the full difficulty distribution, source coverage varies
across sub-disciplines, and expert-authored predicates may reflect the available
notation and verification conventions. Results may also change with prompts, providers, sampling parameters, renderer versions, tool permissions, and repair
budgets. Accordingly, conclusions should be restricted to the documented task
set and Harnesses, and broader generalization requires additional sources,
independent expert calibration, and replication under new configurations.

Agent and direct baselines are independent experimental conditions, so
\Cref{tab:agent-cases} provides challenge examples rather than evidence that an
agent will inherit a direct baseline's outcome. Automatic validators can also
miss mathematically equivalent forms or accept visually legible but pedagogically
misleading artifacts. Educational and research users should therefore verify
figures against the underlying propositions, especially for unfamiliar notation,
new theorems, or settings where an incorrect diagram could materially affect a
proof or explanation.

\section{Conclusion}
\label{sec:conclusion}

This paper proposes MathVis, a benchmark for generating mathematical figures as executable TikZ/SVG programs. Its core design separates technical validity, mathematical predicates, graded presentation scores, and full-specification strict delivery through Gold IR and atomic rubrics. On the test set, the observed strict $D$-PSR values of the 16 evaluated configurations range from $9.18\%$ to $30.73\%$. Future work should expand
independent expert calibration, complete model-family and agent runs, and report
efficiency trajectories.

\subsection*{AI Usage Declaration}

Generative AI tools were used to assist in literature discovery, paper drafting and language editing, LaTeX typesetting, benchmark document organization, and some dataset and evaluation organization work. The authors independently verified the statistics, mathematical definitions, citations, and final artifacts reported in this paper.
The authors reviewed all AI-assisted text, citations, statistics, and artifacts and take responsibility for the final paper, dataset, code, and all statements.

\subsection*{Ethics Statement}

This research does not involve human subjects or personal data. Benchmark materials record sources and rights statuses and are redistributed to the extent permitted by the relevant licenses. Passing the automatic evaluation is not sufficient to ensure that a figure is suitable for classroom or research use; teachers and mathematical experts should still verify figures according to the underlying propositions. The current paper documents the calibration protocol and review workflow, but it does not report a finalized inter-rater summary table.

\subsection*{Reproducibility Statement}

The main text defines the task format, Gold IR, technical and semantic gates, strict delivery indicator, and the six metrics reported in the scorecard. The current artifact contains question-level logs for three frozen configurations and aggregate scorecard rows for the other configurations. It therefore supports output-level recomputation only for those three frozen configurations; the remaining rows can be checked for consistency with their stated denominators but cannot be independently reconstructed from this artifact.

\subsection*{Artifact Inventory}

Analysis artifacts include frozen strict run and rubric JSONL files, Gold IR records, question-level and model-level TSV summaries, official question lists, and XLSX scorecards. Identity-related metadata are removed from the released artifact, and the rights status of source figures is preserved in the source records.

\subsection*{Model Family Scorecard Notes}

\Cref{tab:family-scorecard} and \Cref{fig:family-scorecard} report the same 16 evaluated model configurations. Each row identifies a model configuration and uses the metric definitions in \Cref{app:metrics}. Question-level logs are retained for three frozen configurations. The other 13 configurations are represented only by aggregate scorecard rows in the current artifact and therefore do not support question-level or paired reanalysis. Qualitative examples name the full model configuration and, when applicable, distinguish the underlying model from the agent Harness. Provider routes, temperature settings, and output budgets belong to the archived Harness definitions, but they are not fully tabulated for every configuration in this paper. Missing token, latency, cost, tool-call, rendering, and output-level fields remain NA and are not inferred.


\bibliographystyle{unsrtnat}
\bibliography{mathvis}

\appendix
\section{Contributions}
\label{app:contributions}

A more detailed, artifact-oriented contribution inventory follows:

\begin{enumerate}
  \item Introduced an evaluation-only benchmark for advanced mathematical figure generation,
        extending coverage beyond predominantly K--12 settings to 27 mathematical sub-disciplines (e.g.,
        differential geometry, complex analysis, linear algebra, analytic
        geometry).
  \item Released a structured task design containing prompts, reference
        materials, Gold IR, atomic rubrics, source information, error codes, and
        acceptable equivalent transformations of figures.
  \item Formalized mathematical reasoning figure generation as constrained
        executable programs, and separately verified the correctness of program
        execution, the correctness of mathematical semantics, and the quality of
        visual expression.
  \item Separated technical validity, mathematical hard predicates,
        soft-constraint diagnostics, and strict deliverability in a unified
        metric definition, while retaining formal definitions of repeated-
        generation and rubric diagnostics for reproducibility.
  \item Reported the observed scorecard metrics for the 16 evaluated model
        configurations under the same 512 official questions and three-rollout
        protocol, with explicit notes on which configurations retain
        question-level logs in the current artifact.
\end{enumerate}

\section{Method-Related Work}
\label{app:method-related-work}

\noindent\textbf{Executable Judges and Automated Verification.} MathGen uses a script-as-a-judge approach to evaluate whether text-to-image models preserve charts, geometric constructions, and symbol layouts in mathematical answers \citep{liu2026mathgen}. Math-Vision Diagrams integrates both text-to-code and text-to-image tasks into a unified evaluation framework for mathematical figure generation, while Diagram-MMU focuses on parsing, editing, and question answering based on existing scientific figures \citep{kashyap2026mathvisiondiagrams, bo2026diagrammmu}. These works indicate that code compilation, scripting, or structured QA can provide verification signals that go beyond simple image similarity. Building on this idea, MathVis uses code executability as a technical gate and Gold IR with atomic predicates as semantic gates.

Following the HumanEval setup for repeated sampling in code generation \citep{chen2021humaneval}, our paper reports single-sample success rates, at least one success statistic (suitable for scenarios with multi-candidate generation and verification), and all success statistics (suitable for strict, direct-use scenarios). We also record data sources and model/Harness configurations in line with established data and model documentation practices \citep{bender2018datastatements, gebru2021datasheets}.

\section{Benchmark Construction Process}
\label{app:construction}

\subsection{Question Screening Process Statistics}
\label{app:screening}

Dataset construction began with 8,114 candidate propositions with potential value for mathematical graphical expression. Candidate questions were
sequentially screened based on source completeness, mathematical visualizability,
instance unique determinability, and validator auditability. The first stage eliminated 432 questions with incomplete source information, retaining 7,682; the second stage eliminated 1,710 questions lacking stable mathematical figure semantics, retaining 5,972; the third stage eliminated 5,345 questions with multiple indistinguishable legal instances, retaining 627; the fourth stage eliminated 115 questions unable to establish sufficient verification evidence, and the process then froze 512 official questions. Each question records only the reason for its first triggered elimination in order to avoid double counting across stages.


This pipeline can be written as the funnel in \Cref{eq:funnel}.

\subsection{Stage 1: Theorem Screening and Instance Specification}
\phantomsection
\label{app:stage1}


Candidate propositions come from textbooks, published research papers, and exercise books with traceable origins. For example, complex analysis, differential geometry, and topology are treated as non-interchangeable task assignments because standard graphical conventions and common errors differ across these three fields. The coverage quantity for each field is determined by the observed density of auditable visual propositions, without forcing an even distribution.

Candidate propositions first undergo a five-level judgment of visualizability. This triage is used to allocate question-construction effort on the candidate pool rather than to summarize the released 512 questions. Levels 1 and 2 are not admitted to the benchmark; among admissible candidates, Levels 3--5 describe increasing stability of auditable visual structure:

\begin{description}
  \item[Level 1] The planar figure only serves a decorative purpose; lines,
        colors, and shapes in the figure have no clear correspondence with
        objects, conditions, or conclusions in the proposition, so it can neither
        assist in explaining the proposition nor be used to judge whether the
        proposition is expressed correctly. Such figures are discarded.
  \item[Level 2] Only schematic diagrams can be drawn to aid
        understanding, but elements in the figure cannot be mapped item by item
        to mathematical objects and relations in the proposition, so the figure
        cannot be used to verify whether the proposition is correctly expressed.
        Such candidates are not admitted to the benchmark.
  \item[Level 3] The proposition has no unique standard illustration and
        requires pre-defined visual conventions such as colors, directions,
        projections, or line styles; once conventions are fixed, mathematical
        objects and relations in the figure can still be verified item by item.
  \item[Level 4] The proposition has relatively stable illustration
        methods widely adopted by the mathematical community. Key objects,
        relations, and quantitative properties in the figure can directly
        correspond to explicit mathematical invariants, formulas, or decision
        conditions. Even if different illustrators use different layouts, they will
        retain the same core mathematical structure. For example, an Argument
Principle diagram needs to show closed curves, zeros, poles, and winding
numbers, which can be verified through point-in-region judgments,
zero/pole counting, and winding number calculations.
  \item[Level 5] The proposition not only has a stable typical
        construction but can also form multiple controlled variants by changing
        parameters, functions, regions, or representations. Each variant
        maintains the same core theorem structure while producing different
        visual results. Such tasks can test whether the model understands
        the proposition rather than memorizing a specific standard figure. For
        example, in a Rouch\'e's Theorem task, functions or circle radii can be
        changed; as long as the new parameters still satisfy the boundary
        inequality, the zero count and figure structure are updated accordingly.
\end{description}

Question constructors then fix a specific instance $\eta_i$. General statements
about regions, surfaces, bundles, or knots are translated into explicit objects,
parameters, directions, and conventions. This instantiation step is applied only when the source and accepted conventions support a canonical benchmarkable instance. Candidates that still admit multiple materially different legal realizations without such a source-supported canonicalization are excluded at Stage 3 rather than being resolved by an arbitrary post-hoc choice.
If the same proposition allows meaningful parameter or representation changes, a standard instance can be accompanied by one or two controlled variants. Each variant is assigned an independent question number. All variants originating from the same proposition or standard instance share the same task family number for internal bookkeeping. This paper releases an evaluation-only set with hidden evaluator assets and does not define or report family-based train/dev/test splits.

Constructors write public specifications and hidden specifications separately.
Public prompts explain the mathematical environment and required artifacts but do not enumerate evaluation checklists. Hidden specifications decompose the instantiated proposition into atomic constraints, map constraints to Gold IR fields, declare hard or soft types, freeze soft constraint weights, record acceptable equivalent forms, and specify allowed evidence. Constraint types and weights must be frozen before any benchmark model is run.


\subsection{Stage 2: Candidate Reference Figure Generation and Automatic Pre-check}
\phantomsection
\label{app:stage2}

The reference figure generation path is determined by the mathematical content.
Functions and complex analysis tasks can use deterministic sampling, contour
calculation, or domain coloring tools. Parametric geometry tasks can use
numerical surface and projection programs. Topology and diagrammatic tasks may
require verified off-the-shelf libraries or manually written vector programs.
Each path exports a complete TikZ or SVG artifact and records the renderer and
numerical settings. Unified canvases, fonts, line widths, and coordinate
conventions are used to reduce visual shortcuts formed by rendering paths. When
complete unification is impossible, path labels are retained for stratified
analysis.

Model-assisted drafting can speed up reference artifact writing but cannot define ground truth. The pipeline records the draft model, prompt, and original output.
Constructors modify or regenerate the artifacts to align them with the source proposition and Gold IR before submitting to Stage 3. When conditions permit, the draft model is kept separate from the formal evaluation configuration. Whether a question is admitted is still determined by expert verification.

Automatic pre-checking forms the first layer of the verification funnel. General
checks verify schema completeness, independent compilation or rendering, resource
bounds, and non-empty outputs. Sub-discipline checks calculate quantities that can be judged at low cost, including contour membership, zero and pole positions, directions, conformal non-degeneracy on sampling grids, metric consistency, or curvature values. Passing the automatic pre-check does not mean the question is admitted. Pre-checks are used to eliminate data entry and execution errors and to attach quantitative evidence so that expert time can focus on semantic judgments. Questions lacking reliable automatic checks proceed to expert review without receiving a default pass value.

\subsection{Stage 3: Expert Verification and Arbitration}
\phantomsection
\label{app:stage3}

Stage 3 distinguishes between task package admission and model output scoring.
Sub-discipline experts first verify source tracking, instance validity, public prompts, reference programs, rendering results, Gold IR, hidden constraints, hard and soft types, weights, and allowed equivalent forms. Every decision is recorded down to the predicate level. When prompts are ambiguous, no source-supported canonical instance can be frozen, reference artifacts violate propositions, IR lacks necessary relations, or constraints reward a source-specific layout rather than mathematically equivalent forms, the task is sent back for revision or removed from the benchmark candidate pool.

The full release protocol designates a 5\% calibration subset stratified by source, sub-discipline, rendering path, and difficulty for independent dual-person review. When there is a disagreement on whether a task can be admitted or on any hard constraint, it is arbitrated by a third qualified expert. Binary correctness is intended to be summarized by raw agreement rates and Cohen's $\kappa$ on the dual-reviewed subset. Ordinal soft judgments use a frozen $\{0, 0.5, 1\}$ scale and are intended to be summarized by weighted agreement rates together with intra-class correlation coefficients or Krippendorff's $\alpha$. The present paper documents this protocol but does not report final aggregated calibration coefficients.

Initially, 8,114 questions were collected. The four recorded elimination stages retain 512 questions in the final set. The third elimination stage refers to candidate propositions for which no source-supported canonical benchmark instance could be frozen after revision; it does not mean that every proposition with more than one legal drawing is automatically excluded. The funnel can be written as

\begin{equation}
  8114 \xrightarrow[\text{Incomplete Source}]{-432}
  7682 \xrightarrow[\text{Math Not Visualizable}]{-1710}
  5972 \xrightarrow[\text{No Canonical Instance}]{-5345}
  627  \xrightarrow[\text{Insufficient Validator}]{-115}
  512.
  \label{eq:funnel}
\end{equation}

Every return in Stage 3 records structured reasons and target stages, including source recovery, instance revision, prompt revision, reference figure repair, IR completion, constraint revision, and renderer repair. The revision ledger records rounds and actual expert time, without inferring work hours from file metadata. The current paper records the existence of these audit fields but does not report aggregate expert hours, return rates, or inter-rater statistics.

\begin{table}[htbp]
  \caption*{Professional work, risks, and archived records at each stage.}
  \footnotesize
  \begin{tabularx}{\linewidth}{@{}lYYY@{}}
    \toprule
    \sotaheader{Stage} & \sotaheader{Professional work} &
    \sotaheader{Main risks} & \sotaheader{Archived records} \\
    \midrule
    Proposition screening &
      Recover visualizable propositions from traceable sources and assign to
      corresponding math sub-disciplines &
      Forcing illustrations for theorems lacking stable 2D semantics &
      Source page, local context, rights status, sub-discipline, and
      visualizability rationale \\
    Instance specification &
      Fix regions, functions, parameters, manifolds, knots, directions, and
      conventions &
      Multiple legal instances having different visual meanings &
      Specific instance $\eta_i$, public prompt $x_i$, allowed equivalent forms,
      and prohibited conditions \\
    Reference figure generation &
      Generate executable candidate figures using numerical, parametric,
      off-the-shelf libraries, or manual vector paths &
      Renderer style becoming a shortcut to identify questions &
      Source program, renderer version, path label, rendering results, and
      editing records \\
    Automatic pre-check &
      Check schema, compilation, non-empty rendering, numerical invariants, and
      mechanically verifiable IR relations &
      Mistaking successful rendering as evidence of mathematical correctness &
      Machine logs, measured quantities, failed check items, and evidence
      confidence \\
    Expert verification &
      Verify source consistency, instance validity, Gold IR coverage, constraint
      types, reference figure correctness, and ambiguous outputs &
      Figure appearance reasonably masking subtle semantic errors &
      Predicate-level decisions, evidence, reviewer sub-discipline, return
      reasons, and arbitration records \\
    \bottomrule
  \end{tabularx}
\end{table}

Each question contains a self-contained prompt and output contract, a reference
figure and an executable TikZ/SVG program, Gold IR, atomic predicates belonging to
five dimensions, and error codes with evidence priority. IR focuses on expressing relations: for example, a coordinate chart task requires not only two ellipses
that look similar, but also intersecting coordinate domains, a specified
intersection, coordinate images in $\mathbb{R}^m$, and mutually inverse
transformation mappings between coordinate images.

Predicates concretize the ``understanding to action'' chain as: Language Parsing
$\rightarrow$ Relation Extraction $\rightarrow$ Implicit Condition Inference
$\rightarrow$ Spatial Planning and Calculation $\rightarrow$ Program Synthesis
$\rightarrow$ Rendering $\rightarrow$ Verification. The evaluation allows alternative layouts but does not allow visually similar approximations to replace the mathematical propositions. Error codes locate object omissions, topological errors, missing or reversed relations, formula mapping errors, label misbindings, incomplete content, and unreadable deliveries, turning a single leaderboard score into an interpretable error profile.

The current version uses automatic auditing and escalates cases with weak automatic evidence or high downstream impact for expert review, with an independent dual-expert calibration protocol. Aggregate calibration statistics are not reported in this version.

\section{Complex Analysis Predicate Family}
\label{app:predicates}

Each question instantiates only the predicates that are mathematically applicable.

\setcounter{table}{2}
\begin{table}[htbp]
  \caption{Complex analysis predicate family.}
  \label{tab:ca-predicates}
  \footnotesize
  \begin{tabularx}{\linewidth}{@{}lYYY@{}}
    \toprule
    \sotaheader{Predicate family} & \sotaheader{Typical hard judgment} &
    \sotaheader{Typical soft judgment} & \sotaheader{Priority evidence} \\
    \midrule
    Regions and boundaries &
      Do regions, holes, boundary branches, inclusion relations, and
      interior/exterior markers have correct topology? &
      Are different branches easy to distinguish? &
      Vector paths, point-in-region tests, and expert review \\
    Directions and traversals &
      Are contour directions mathematically correct and consistent with arrow or
      parameter directions? &
      Are direction hints clear and not overly crowded? &
      Path ordering, arrow extraction, and local tangent checks \\
    Functions and images &
      Are pre-image domains, image domains, branch selections, and boundary
      correspondences correct? &
      Is the visual correspondence between pre-image and image clear? &
      Gold IR edge matching, function sample values, and label recognition \\
    Zeros, poles, and critical points &
      Are multiplicities, positions relative to contours, and point type symbols
      correct? &
      Are different point types easy to distinguish? &
      Numerical roots or pole checks and region membership judgments \\
    Quantitative invariants &
      Are modulus, argument, winding number, residue relations, or inequalities
      on specified sets correct? &
      Do invariants establish clear visual connections with relevant objects? &
      Symbol or numerical checks and formula parsing \\
    Branches and branch cuts &
      Are branch cuts, Riemann surfaces, discontinuities, and analytic
      continuation paths mathematically compatible? &
      Are different layers or sides of branch cuts easy to distinguish? &
      Topological adjacency, sampled arguments, and expert review \\
    Labels and formulas &
      Is each formula bound to the correct object without altering signs, ranges,
      or function names? &
      Are non-critical labels clear and non-overlapping? &
      Structured text extraction and spatial binding \\
    \bottomrule
  \end{tabularx}
\end{table}

\section{Expert Verification Rubric Template}
\label{app:rubric}

Stage 3 uses two related forms. The task admission form verifies the benchmark question itself. The model output scoring form is only used to evaluate
model artifacts after the task package is frozen. This separation helps avoid misjudging errors in reference programs or Gold IR as model failures.

\begin{table}[htbp]
  \caption{Task admission rubric.}
  \label{tab:admission}
  \small
  \begin{tabularx}{\linewidth}{@{}lYY@{}}
    \toprule
    \sotaheader{Field} & \sotaheader{Admission decision question} &
    \sotaheader{Required evidence} \\
    \midrule
    Source consistency &
      Does the instantiated question maintain the source proposition, notation,
      assumptions, and directions? &
      Source page or original figure, local context, and constructor's
      verification notes \\
    Instance validity &
      Are regions, functions, parameters, objects, and conventions clear and
      mutually consistent? &
      Symbol table, parameter check, and executable or numerical witness \\
    Prompt sufficiency &
      Can a qualified solver reconstruct the required artifact without knowing
      the evaluation-side information? &
      Independent reconstruction record based solely on the prompt \\
    Gold IR coverage &
      Does the IR contain all objects, relations, formulas, invariants,
      equivalent forms, and prohibited conditions required by the instance? &
      Item-by-item correspondence from source proposition to IR fields \\
    Constraint validity &
      Is each atomic constraint decidable, non-redundant, linked to IR evidence,
      and correctly classified as hard or soft? &
      Constraint card with type rationale, weight rationale, and failure codes \\
    Reference artifact correctness &
      Does the executable reference artifact satisfy all hard constraints and
      expected soft specifications? &
      Compilation logs, rendering results, automatic measurements, and expert
      sign-off \\
    Equivalent form policy &
      Does the question accept mathematically equivalent layouts while rejecting
      transformations that alter the proposition? &
      Positive and negative examples of allowed transformations \\
    \bottomrule
  \end{tabularx}
\end{table}

Each output predicate card contains the fields id, Gold-IR path, decision question, type, weight, allowed equivalence, evidence method, satisfaction, error code, and reviewer confidence. Hard constraints use $h \in \{0, 1\}$ and have no
compensatory weights. Soft constraints use $s \in \{0, 0.5, 1\}$, with their
positive weights normalized within the question. Experts can choose \textbf{insufficient evidence}, which triggers an escalated review and cannot be converted to a pass.

\section{Metric Definitions}
\label{app:metrics}

\subsection{Task Definition and Output Contract}

Each sample is represented as

\begin{equation}
  \mathcal{T}_i = (x_i, r_i, g_i, a_i, e_i),
  \label{eq:task}
\end{equation}

where $x_i$ is a self-contained natural language description, $r_i$ is the
reference figure and executable reference program, $g_i$ is the Gold IR, $a_i$
is the list of atomic rubrics, and $e_i$ is the error code and evidence
specification. The model can only see $x_i$ and the output contract, and must
generate a complete, independent TikZ document or a directly renderable SVG
document. The reference figure and reference program are held out from the model and used only for evaluation.

The benchmark follows a problem-driven pipeline: Source Problem
$\rightarrow$ Self-contained Prompt and Gold IR $\rightarrow$ Construction
Planning and Executable Code $\rightarrow$ Compilation and Rendering
$\rightarrow$ Technical, Structural, Visual, and Expert Verification.
Verification feedback is reserved for separately budgeted repair experiments; direct baselines do not receive feedback. Therefore, MathVis measures not image
imitation, but the ability to translate problem understanding into executable,
verifiable, and editable artifacts.

This paper defines mathematical figures as 2D visual representations in which objects, spatial relations, labels, and formulas encode propositions that can be verified against structured specifications. As long as mathematical relations remain unchanged, global translation, uniform scaling, and explicitly permitted line-style changes are acceptable.

``Mathematical reasoning'' in this paper does not refer to the unobservable
internal chain of thought of the model, nor does it require the model to submit
a Chain-of-Thought. We operationalize it as: can the model extract mathematical
objects and relations from natural language, synchronously update the graphic
structure when parameters or relations undergo controlled changes, and maintain
the consistency of unchanged constraints? This operational definition is implemented through Gold-IR comparison and explicit/implicit constraint checks in the reported evaluation.

Let the model be $m$ and the number of questions be $N=512$. Each question has
three rollouts, giving $3N=1536$ model outputs. For each rollout, define the
technical gate

\begin{equation}
\begin{aligned}
T_{i,m,r}=\mathbb{1}\!\big[&\text{Format Correct}
  \land \text{Parsable or Compilable}\\
  &\land \text{Rendering Successful}
  \land \text{Figure Non-empty}\big].
\end{aligned}
\label{eq:t-gate}
\end{equation}

Let $H_{\mathrm{math},i,m,r}\in\{0,1\}$ denote whether all mandatory
mathematical hard predicates pass. Let each atomic soft constraint take a value
in $\{0,0.5,1\}$, and let $S_{i,m,r}\in[0,1]$ be the normalized weighted
aggregate. The strict deliverability indicator is

\begin{equation}
D_{i,m,r}=T_{i,m,r}\,H_{\mathrm{math},i,m,r}\,
\mathbb{1}[S_{i,m,r}=1].
\label{eq:d-gate}
\end{equation}

This definition keeps technical and mathematical failures non-compensatory while
preserving partial soft scores for diagnostics. It implies
$D_{i,m,r}\leq T_{i,m,r}$ and $D_{i,m,r}\leq S_{i,m,r}$. The current workbook
supports aggregate reporting of $T$, $S$-PSR, and $D$-PSR. It does not contain
the output-level predicate records required to report a separate aggregate
$H_{\mathrm{math}}$ rate.

\subsection{Technical Gate $T$}

The observed technical gate pass rate is

\begin{equation}
  T_m = \frac{100}{3N} \sum_{i=1}^{N} \sum_{r=1}^{3} T_{i,m,r}.
  \label{eq:t}
\end{equation}

Here $k_T=\sum_{i=1}^{N}\sum_{r=1}^{3}T_{i,m,r}$ is the number of outputs
that pass all four technical checks, so the reported value is $100k_T/1536$.

\subsection{\texorpdfstring{$S$-PSR}{S-PSR}}

The soft-constraint per-sample rate is defined as

\begin{equation}
  S\text{-PSR}_m = \frac{100}{3N} \sum_{i=1}^{N} \sum_{r=1}^{3} S_{i,m,r}.
  \label{eq:spsr}
\end{equation}

This diagnostic averages the aggregate soft-constraint score
$S_{i,m,r}\in[0,1]$ and thereby preserves partial credit across atomic soft
constraints. It does not override the technical or mathematical hard gates. The table reports the observed rate computed over all questions and rollouts for each model. By construction, $S$-PSR is no lower than $D$-PSR.

\subsection{\texorpdfstring{$D$-PSR}{D-PSR}}

The observed strict deliverability rate is defined as

\begin{equation}
  D\text{-PSR}_m = \frac{100}{3N} \sum_{i=1}^{N} \sum_{r=1}^{3} D_{i,m,r}.
  \label{eq:dpsr}
\end{equation}

It represents the proportion of outputs that satisfy the technical gate,
all mandatory mathematical hard predicates, and complete soft-constraint
satisfaction. With $k_D=\sum_{i=1}^{N}\sum_{r=1}^{3}D_{i,m,r}$, the reported
value is $100k_D/1536$.

\subsection{Rubric Coverage}

Let the $i$-th question contain $M_i$ atomic rubrics, let
$\rho_{i,m,r,j} \in [0,1]$ represent the satisfaction degree of the $j$-th
rubric, and let $w_{ij}$ be its importance weight. The question-level rubric
score is

\begin{equation}
  R_{i,m,r} = \frac{\sum_{j=1}^{M_i} w_{ij}\, \rho_{i,m,r,j}}{\sum_{j=1}^{M_i} w_{ij}}.
  \label{eq:rubric}
\end{equation}

Converting this to a score from 0 to 10:

\begin{equation}
  \operatorname{Rubric}_m = \frac{10}{N} \sum_{i=1}^{N} \frac{1}{n_i} \sum_{r=1}^{n_i} R_{i,m,r}.
  \label{eq:rubric10}
\end{equation}

If binary rubrics are used, then $\rho_{i,m,r,j} \in \{0, 1\}$, where $1$ means
the item is fully satisfied, and $0$ means it is not satisfied or evidence is
insufficient.

\subsection{\texorpdfstring{\passatthree{}}{pass@3}}

Generate three independent outputs for each question. As long as at least one output strictly passes, the question is counted as successful. Here $N=512$, so the denominator is the number of questions rather than the number of rollouts.

\begin{equation}
  \text{\passatthree}_m
    = \frac{100}{N} \sum_{i=1}^{N}
      \mathbb{1}\!\left[\max_{r \in \{1,2,3\}} D_{i,m,r} = 1\right]
    = \frac{100}{N} \sum_{i=1}^{N}
      \mathbb{1}\!\left[D_{i,m,1} \lor D_{i,m,2} \lor D_{i,m,3}\right].
  \label{eq:passat3}
\end{equation}

This metric is suitable for settings in which multiple candidate results can be generated, and the best result is selected by a validator, teacher, or agent.

\subsection{\texorpdfstring{\passthree{}}{pass3}} Generate three independent outputs for each question. A
question is counted as successful only if all three strictly pass. Here $N=512$, so the denominator is the number of questions rather than the number of rollouts:

\begin{equation}
  \text{\passthree}_m
    = \frac{100}{N} \sum_{i=1}^{N}
      \mathbb{1}\!\left[\min_{r \in \{1,2,3\}} D_{i,m,r} = 1\right]
    = \frac{100}{N} \sum_{i=1}^{N} \prod_{r=1}^{3} D_{i,m,r}.
  \label{eq:passall3}
\end{equation}

This metric measures the stability of repeated generation: as long as at least one of the
three rollouts fails, the question is scored as $0$ in \passthree{}.

We consider two application settings rather than making causal claims about downstream user outcomes. Education is the primary application setting, including textbooks, lecture notes, assignments, and tutoring materials; incorrect arrows or labels may directly mislead learners. Research communication is a secondary setting, including research notes, paper sketches, and theoretical communication in fields such as topology, differential geometry, algebra, and analysis, where higher-order relations and notation must remain auditable. The published metadata records these settings separately; for educational use, we recommend the full rubric hard gate. The current paper does not claim that benchmark scores directly predict teacher or student outcomes; such validity still requires separate educational studies.

\section{Result Statistics}
\label{app:results}

\subsection{Model Scorecard}
\label{app:scorecard}

\setcounter{table}{5}
\begin{table*}[htbp]
  \caption{Scorecard for the 16 evaluated model configurations on the 512 official questions. All rates are observed percentages. \passone{}, \passatthree{}, and \passthree{} each use 512 question-level outcomes; $D$-PSR, $T$, and $S$-PSR use $512\times3=1536$ rollout-level outcomes. \passatthree{} measures whether at least one of three outputs strictly passes, whereas \passthree{} measures whether all three strictly pass.}
  \label{tab:family-scorecard}
  \centering
  \small
  \setlength{\tabcolsep}{6pt}
  \begin{tabular}{@{}lNNNNNN@{}}
    \toprule
    \sotaheader{Model} &
    \sotaheader{\passone} &
    \sotaheader{$D$-PSR} &
    \sotaheader{$T$} &
    \sotaheader{$S$-PSR} &
    \sotaheader{\passatthree} &
    \sotaheader{\passthree} \\
    \midrule
    GPT-6-Astra & 31.25 & 30.73 & 94.40 & 31.20 & 38.28 & 24.22 \\
    Claude Fable 5 & 26.56 & 25.13 & 84.11 & 26.07 & 29.30 & 19.14 \\
    Gemini 3.1 Pro Preview & 22.85 & 20.70 & 78.13 & 20.70 & 27.34 & 15.23 \\
    GPT-5.6-sol & 21.48 & 19.99 & 92.32 & 20.05 & 24.22 & 16.60 \\
    Claude Opus 4.8 & 20.90 & 19.92 & 78.91 & 20.12 & 24.80 & 15.04 \\
    Doubao Seed 2.1 Pro & 20.90 & 19.14 & 81.05 & 20.03 & 24.22 & 14.84 \\
    GLM 5.2 (Direct) & 18.95 & 17.12 & 82.23 & 18.07 & 22.27 & 13.09 \\
    Doubao Seed 2.0 Pro & 18.16 & 16.15 & 77.54 & 17.12 & 21.29 & 12.30 \\
    GLM 5.2 (OpenRouter) & 17.58 & 16.08 & 78.13 & 17.06 & 21.09 & 12.11 \\
    Claude Sonnet 4.6 & 16.80 & 15.56 & 79.88 & 16.47 & 20.12 & 11.91 \\
    DeepSeek V4 Flash 0731 & 17.19 & 15.04 & 78.06 & 16.12 & 20.31 & 11.13 \\
    Doubao Seed 2.1 Turbo & 16.21 & 14.58 & 76.56 & 15.47 & 19.14 & 9.57 \\
    Gemini 3.5 Flash & 16.02 & 14.06 & 75.13 & 15.11 & 18.95 & 10.74 \\
    Hunyuan HY3 & 15.82 & 13.67 & 76.11 & 14.53 & 18.16 & 10.16 \\
    Doubao Seed 2.0 Lite & 11.33 & 10.09 & 72.07 & 11.08 & 13.28 & 7.23 \\
    Claude Haiku 4.5 & 10.35 & 9.18 & 71.55 & 10.06 & 12.30 & 6.05 \\
    \bottomrule
  \end{tabular}
\end{table*}
\FloatBarrier

\subsection{Additional Result Analysis}
\label{app:result-analysis}

\subsubsection{Metric Separation}

The five predicate groups---objects and structures, relations and quantities,
labels and formulas, completeness and consistency, and readability and
layout---remain useful for error diagnosis. We do not report aggregate
``full satisfaction rates'' for these groups in this version because the
retained workbook does not contain the output-level predicate records needed
to reproduce those rates. Technical validity is reported separately as $T$;
its values range from $71.55\%$ to $94.40\%$ and must not be interpreted as
mathematical hard-predicate accuracy.

\subsubsection{Questions and Observed Scores}

The questions in the benchmark provide unified prompts, renderers, Gold IR, and delivery gates while maintaining source and mathematical sub-discipline coverage. On this question set, the observed strict $D$-PSR values across the 16 evaluated configurations range from $9.18\%$ to $30.73\%$.

\subsubsection{Qualitative Error Analysis}

Gold IR enables errors to be checked at the mathematical proposition level.
Common errors include: \textbf{relation errors}, such as missing or reversed
arrows, inclusions, associations, directions, or inverse mappings;
\textbf{mapping and formula errors}, such as incorrect signs, value ranges,
branches, or label bindings; \textbf{completeness errors}, meaning some objects
exist but the conjunction of required facts is incomplete; and \textbf{technical
errors}, such as blank outputs, incomplete source code, compilation failures, or
rendering failures.

\setcounter{table}{1}
\begin{table}[H]
  \caption{Reproducibility checklist: fields recorded for evaluation.}
  \label{tab:reproducibility}
  \small
  \begin{tabularx}{\linewidth}{@{}lYY@{}}
    \toprule
    \sotaheader{Field} & \sotaheader{Definition} & \sotaheader{Current status} \\
    \midrule
    Questions & Question ID, source, prompt, reference materials, and Gold IR &
      512 official questions \\
    Output contract & Complete independent TikZ or SVG, output code only & Executed \\
    Technical gate & Parsing / compilation / rendering / non-empty output & Audited \\
    Semantic gate & Rubrics & Diagnostic report \\
    Delivery gate & Technical gate, rubrics, readability & Main strict results \\
    Repeated runs & Outputs per question under the same prompt and Harness &
      All 16 configurations: 3 reported rollouts; question-level logs retained for 3 configurations \\
    Statistics & Point estimates and uncertainty intervals &
      Observed point estimates reported; bootstrap intervals not reported in this version \\
    Sampling configuration & Provider route, temperature, and output budget &
      Partially documented in Harness metadata; not tabulated for all 16 configurations in this artifact \\
    Family coverage & Model ID, provider, repetition count &
      Listed in unified scorecard \\
    Calibration statistics & Agreement coefficients, labor, and return rates &
      Protocol defined; aggregate statistics not reported in this version \\
    Agent auditing & Planning, compilation, verification, and repair trajectories &
      Protocol defined; recorded separately from direct baseline \\
    Efficiency & Tokens, latency, cost, calls, and rendering counts &
      Rounds $=1$, calls $=0$; rest NA \\
    Test governance & Held-out references, programs, and hidden constraints & Implemented \\
    \bottomrule
  \end{tabularx}
\end{table}

\end{document}